\documentclass[12pt]{article}

\usepackage{newtxtext,newtxmath}

\usepackage{graphicx}

\usepackage[letterpaper,margin=1in]{geometry}

\renewenvironment{abstract}
	{\quotation}
	{\endquotation}

\date{}

\makeatletter
\renewcommand{\fnum@figure}{\textbf{Figure \thefigure}}
\renewcommand{\fnum@table}{\textbf{Table \thetable}}
\makeatother

\usepackage{scicite}

\usepackage{url}

\usepackage{booktabs}

\def\scititle{
	BeyondMimic: From Motion Tracking to Versatile Humanoid Control via Guided Diffusion
}
\title{\bfseries \boldmath \scititle}

\author{
	Qiayuan~Liao$^{1\ast\dagger}$,
	Takara~E.~Truong$^{2\dagger}$,
	Xiaoyu~Huang$^{1\dagger}$, \\
    Yuman Gao$^{1}$,
	Guy~Tevet$^{2}$,
	Koushil~Sreenath$^{1\ddagger}$,
	C.~Karen~Liu$^{2\ddagger}$\and
	\small$^{1}$University of California, Berkeley, CA 94720, USA.\and
	\small$^{2}$Stanford University, Stanford, CA 94305, USA.\and
	\small$^\ast$Corresponding author. Email: qiayuanl@berkeley.edu\and
	\small$^\dagger$These authors contributed equally to this work; order decided by coin toss.\and
	\small$^\ddagger$Equal advising; order mirrors the coin toss, with the other lab listed last.
}

\usepackage{xcolor}

\begin{document} 

\maketitle

\begin{abstract} \bfseries \boldmath
The human-like form of humanoid robots positions them uniquely to achieve the agility and versatility in motor skills that humans possess. Learning from human demonstrations offers a scalable approach to acquiring these capabilities. However, prior works either produce unnatural motions or rely on motion-specific tuning to achieve satisfactory naturalness. Furthermore, these methods are often motion- or goal-specific, lacking the versatility to compose diverse skills, especially when solving unseen tasks. We present BeyondMimic, a framework that scales to diverse motions and carries the versatility to compose them seamlessly in tackling unseen downstream tasks.
At heart, a compact motion-tracking formulation enables mastering a wide range of radically agile behaviors, including aerial cartwheels, spin-kicks, flip-kicks, and sprinting, with a single setup and shared hyperparameters, all while achieving state-of-the-art human-like performance. 
Moving beyond the mere imitation of existing motions, we propose a unified latent diffusion model that empowers versatile goal specification, seamless task switching, and dynamic composition of these agile behaviors. Leveraging classifier guidance, a diffusion-specific technique for test-time optimization toward novel objectives, our model extends its capability to solve downstream tasks never encountered during training, including motion inpainting, joystick teleoperation, and obstacle avoidance, and transfers these skills zero-shot to real hardware.
This work opens new frontiers for humanoid robots by pushing the limits of scalable human-like motor skill acquisition from human motion and advancing seamless motion synthesis that achieves generalization and versatility beyond training setups.

\end{abstract}

\subsection*{Introduction}

Humans have long envisioned a future where humanoid robots live and work alongside us in our everyday environments. These humanoids would possess the versatility to perform diverse tasks in the environment designed for humans like Data from Star Trek: The Next Generation (1987-1994), the agility to move in complex worlds with ease like Astro from Astro Boy (2009), and the ability to collaborate with humans through natural, gentle interactions like Baymax from Big Hero 6 (2014). Achieving this vision requires endowing humanoid robots with the intelligence to move and interact as we do, expressing intent through body language, operating fluidly in human environments, and coordinating seamlessly in shared tasks. In this work, we take a step toward realizing that vision.

Humanoid robots share a similar morphology with humans, presenting a strong opportunity to acquire the same skills we have. Learning from human demonstrations provides a scalable way to develop these skills while naturally capturing human-level agility and human-like behaviors. Yet coordinating their dozens of joints and actuators poses significant challenges due to the high dimensionality of the control problem. A floating base adds another layer of complexity, as stable balancing itself remains difficult to achieve, let alone natural, human-like movement. In contrast, humans not only acquire agile skills such as sports and dancing, but are also highly versatile: we can transition through numerous motor skills as tasks demand, and solve simple unseen tasks by composing skills we have had. Endowing these robots with the same agility, naturalness, and versatility remains a grand-challenge in robotics.

The pursuit of humanoid control has historically advanced through model-based paradigms that pair simplified dynamics with hierarchical control~\cite{kuindersma2016optimization, wensing2023optimization}. 
In practice, a few low-frequency planners generate future trajectories over coarse variables such as center of mass (CoM), momentum, contact schedule, and foot placement~\cite{kajita2003biped, pratt2006capture, deits2014footstep, dai2014whole, hereid2015hybrid, koenemann2015whole, kuindersma2016optimization}. A high-frequency, low-level controller then tracks these references with a higher-fidelity model in a myopic manner~\cite{khatib2003unified, herzog2016momentum, wensing2013generation, kuindersma2016optimization, khazoom2022humanoid}. This split keeps computation tractable, but the simplifications impose limits. Kinematic and dynamic surrogates often yield unnatural motions---such as constant CoM height or persistently bent knees~\cite{li2013quasi, carpentier2017learning, fasano2024efficient}---that restrict robots to a small subset of their full motion range. Attempts to improve motion quality by shaping the momentum cost introduce some arm swing and upper-body rotation but remain limited to local stylistic adjustments without restoring overall human-like coordination~\cite{wensing2016improved, chen2023integrable}. 
In addition, transferring control stacks to the real world, further introduces significant unmodeled dynamics and modeling errors in highly dynamic motions~\cite{chignoli2021humanoid}, and model-based, contact-explicit planning remains difficult for contact-rich skills such as ground rolling~\cite{subburaman2018online}. 

To overcome these challenges, learning-based approaches, particularly reinforcement learning (RL), have emerged as promising alternatives. With appropriate reward shaping and sim-to-real transfer, RL-trained humanoids can now perform diverse locomotion behaviors, including walking on flat ground~\cite{radosavovic2024real} and slopes~\cite{radosavovic2024learning, liao2025berkeley}, climbing stairs~\cite{long2025learning, siekmann2021blind}, running~\cite{olkin2025chasing}, traversing challenging terrains such as stepping stones or narrow beams~\cite{he2025attention, wang2025beamdojo}, and even basic compliance control~\cite{zhi2025learning}. However, these successes rely on hand-crafted rewards tuned for specific tasks, which must be redesigned for each new behavior. This makes scaling up to a large repertoire of skills prohibitively costly. Moreover, hand-crafted objectives often produce unnatural movements marked by continuous stepping, bent knees, and heavy impacts, since “naturalness” and “human-likeness” cannot be easily expressed as explicit optimization terms. Although targeted fixes such as improving arm swing via centroidal momentum optimization~\cite{lee2025learning} have been proposed, they remain case-specific and fail to generalize across diverse motions.

A promising alternative is to leverage large-scale human motion data to directly learn human-level skills. 
In model-based settings, libraries of human motion trajectories
can serve as effective warm starts for online controllers, yielding encouraging results~\cite{varin2021estimation, deits2023robot}. Yet these methods still struggle with robustness outside laboratory conditions. 
In comparison, RL-based learning from human motion via motion tracking has produced agile and natural skills, but offers limited adaptability at deployment. Specifically, DeepMimic-style reward formulations~\cite{jason_deepMimic} have demonstrated jumping and turning~\cite{he2025asap}, single-leg balancing~\cite{zhang2025hub}, and even martial-arts sequences~\cite{xie2025kungfubot}, but these policies rely on motion-specific tuning and do not generalize beyond the motions seen during training. Improving upon it, Adversarial Motion Priors (AMPs)~\cite{jason_AMP} learn motion ``styles" for task-specific control~\cite{wang2025physhsi} and generalize beyond specific motion clips, but these policies are generally not reusable across tasks and require retraining from scratch. 

To improve versatility, prior work has taken two main routes. The first is hierarchical control, which combines a task-agnostic motion tracker~\cite{ji2024exbody2, ze2025twist, li2025clone, yin2025unitracker} with a task-level motion planner~\cite{he2025omnih2o, fu2024humanplus}. This enables reuse of learned skills for new tasks but typically sacrifices agility and naturalness, and, due to decoupled training, suffers from planner–controller mismatches~\cite{xu2025parc, Diffuse-CLoC}. To avoid this mismatch, the second route uses multi-task generative models to learn the motion distributions directly. Among these, variational autoencoder (VAE)–based models have demonstrated agility and skill composition~\cite{zeng2025behavior, shao2025langwbc}. 
However, their versatility remains limited as they depend on explicit goal conditioning during training and generalize poorly to tasks with implicit or hard-to-specify objectives, such as obstacle avoidance or long-horizon navigation, yielding out-of-distribution, jittery motions and degraded naturalness~\cite{wu2025uniphys}. 

\begin{figure}
    \centering
    \includegraphics[width=1.0\textwidth]{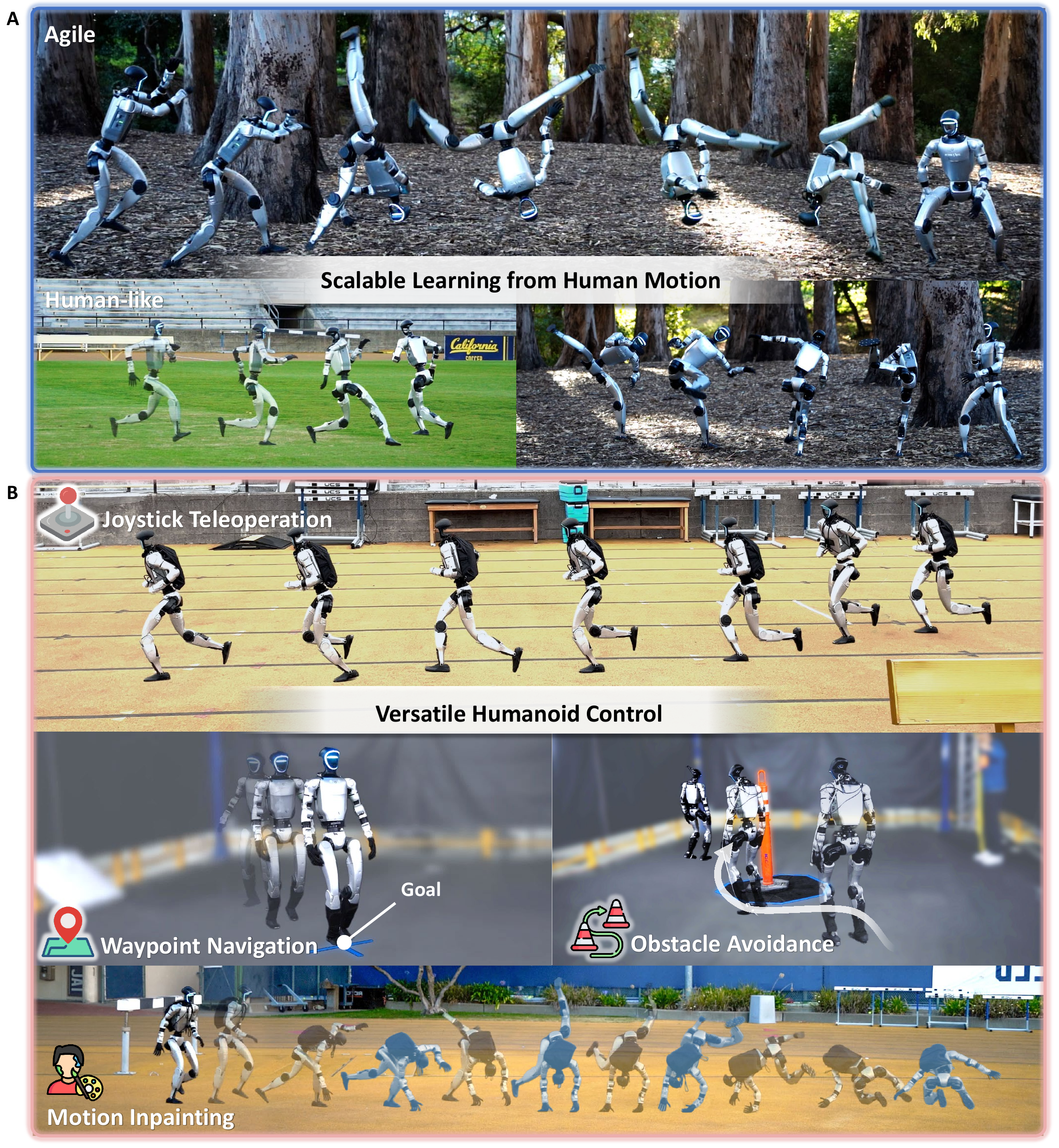}
    \caption{\textbf{Overview of the proposed versatile humanoid control framework.} (\textbf{A}) Scalable and robust learning from human motions with agile, human-like behaviors via motion tracking. (\textbf{B}) Versatile control over unseen downstream tasks with diverse learned motor skills via guided diffusion.}
    \label{fig:teaser}
\end{figure}

Here, we present BeyondMimic, a scalable framework that enables humanoid robots to achieve human-level agility, naturalness, and versatility by learning from diverse, unlabeled human motions and synthesizing them seamlessly to solve unseen real-world tasks via online optimization at inference time. Shown in Fig.~\ref{fig:teaser} and Movie S1, our approach deploys outdoors across various conditions, demonstrating agile and natural motions that surpass prior work in humanoid robotics. Furthermore, we present the first real-world deployment of a flexible unified controller that handles diverse downstream tasks with a variety of skills through online planning alone, without any task-specific training, fine-tuning, or policy optimization.

BeyondMimic is built on two key insights. 
First, we depart from the common practice of introducing complex RL formulation to foster successful sim-to-real transfer. In conventional setups, heavy domain randomization, reward regularization, and complex observations are used to compensate for idealized simulation dynamics during real-world deployment. However, this often degrades performance, necessitating motion-specific tuning. Instead, we demonstrate that scalable, high-quality motion tracking can emerge from a compact, principled formulation. 
To achieve a generalizable formulation without any motion-specific tuning, we carefully model the robot actuation based on classical mechanics principles and ensure proper system implementation to minimize deployment discrepancies, such as delays.
This disciplined design allows us to apply domain randomization only to truly uncertain physical properties, maintaining robustness without diluting the control objective. 
As a result, we are able to distill our reward formulation down to only three regularization terms essential for physically consistent behaviors, in addition to a unified task reward that generalizes across motions. With a simple yet effient formulation, RL training becomes straightforward: we use the same model parameters, reward function, and hyperparameters across a diverse set of skills and different physics engines, allowing human-like behaviors to transfer zero-shot to real humanoids.

Second, versatility must extend beyond training-time diversity. The robot must adapt at test time to novel objectives after deployment. BeyondMimic synthesizes the atomic skills trained by RL into novel sequences for zero-shot, task-specific control at test time. The key concept is that diffusion models, unlike VAEs or AMPs, not only capture the complex, multimodal distribution of diverse skills, but also support online optimization for new objectives at test time. Specifically, they learn the gradient fields of the data distribution~\cite{song2020score} rather than the distribution itself, enabling gradient-based optimization toward arbitrary differential objectives at test time, a technique called \emph{classifier guidance}. To leverage this capability, we employ a state-action co-diffusion model that acts in a predictive control manner, allowing cost formulation on both future states and actions. This distinct design choice enables a single, unified controller to solve a wide range of unseen tasks with unlabeled data, such as waypoint navigation, joystick teleoperation, and obstacle avoidance, while preserving the natural style and dynamic quality of the original human motions. 

In summary, BeyondMimic offers the first unified framework to achieve human-level agility, human-like behaviors, and zero-shot task versatility. By combining scalable human motion learning with diffusion-based online optimization, it bridges the gap between specialized motion tracking and general-purpose task adaptability. This work provides a practical foundation for humanoid robots capable of learning, adapting, and composing skills to move and act seamlessly within human environments.

\section*{Results}
\noindent Our framework successfully learns to control a humanoid robot to perform a wide repertoire of complex, human-like behaviors. 
We organize our results into two parts. First, we demonstrate scalable learning of diverse human motions, highlighting both the diversity of acquired skills and their human-like quality, including agility and naturalness. Second, we show that these learned skills can be flexibly composed and repurposed at inference time. Using diffusion-based online optimization, our controller adapts to new, unseen tasks such as command-conditioned locomotion, scene-aware navigation, and motion completion from keyframes, while enabling smooth transitions between skills without retraining.

\subsection*{Scalable Learning from Human Motions}
Our framework demonstrated scalable learning from diverse human motions. 
With motion tracking, the robot successfully learned a broad set of agile, stylish, human-like skills using a \emph{single} formulation and \emph{shared} hyperparameters, without motion-specific tuning. 

\subsubsection*{Diverse Skills}
\begin{figure}
    \centering
    \includegraphics[width=1.0\textwidth]{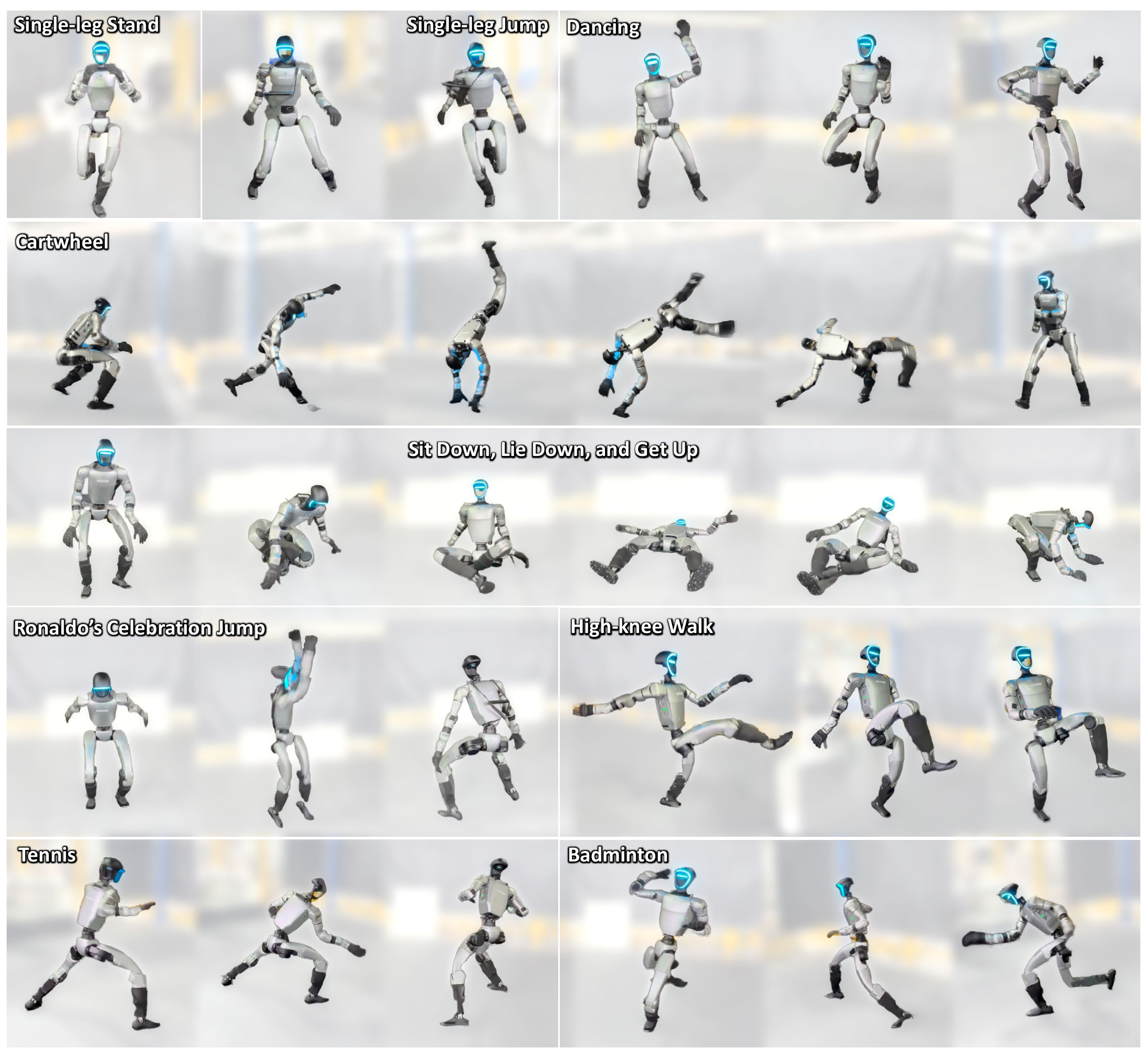}
    \caption{\textbf{Diverse motion tracking policies deployed on a real humanoid robot.}
    \label{fig:more_motion}
    We demonstrate accurate and robust tracking across a wide spectrum of motions, ranging from static to highly dynamic and from athletic feats to stylized motions, showing the scalability of our framework. In total, 30 distinct motion clips are deployed on hardware; with a subset shown here due to space. See Movie~S2 for the full repertoire. }
\end{figure}

In total, we trained on approximately 2.5 hours of diverse human motions and successfully validated all motions in high-fidelity simulation. To assess sim-to-real transfer, we deployed 30 representative clips (totaling 15 minutes) on the physical robot, showing that the skills learned in simulation transferred reliably to hardware (Fig.~\ref{fig:more_motion}).
The demonstrated behaviors spanned static and balance-critical motions such as single-leg standing and standing up varying poses; highly dynamic skills such as single-leg jumps, turn kicks, forward jumps with 180° and 360° spins, and cartwheels; and stylized or expressive behaviors such as elderly-style walking, dance sequences, and sport moves, with recognizable human-like quality. 
Importantly, many challenging clips were trained jointly with multiple other skills, with the entire reference motion exceeding three minutes. Despite this, the policy is able to retain agility and stylistic details, indicating that our unified formulation scales across diverse human motions without motion-specific tuning. The list of validated motions is provided in Movie S2.

\subsubsection*{Human-level Agility}
\begin{figure}
    \centering
    \includegraphics[width=1.0\textwidth]{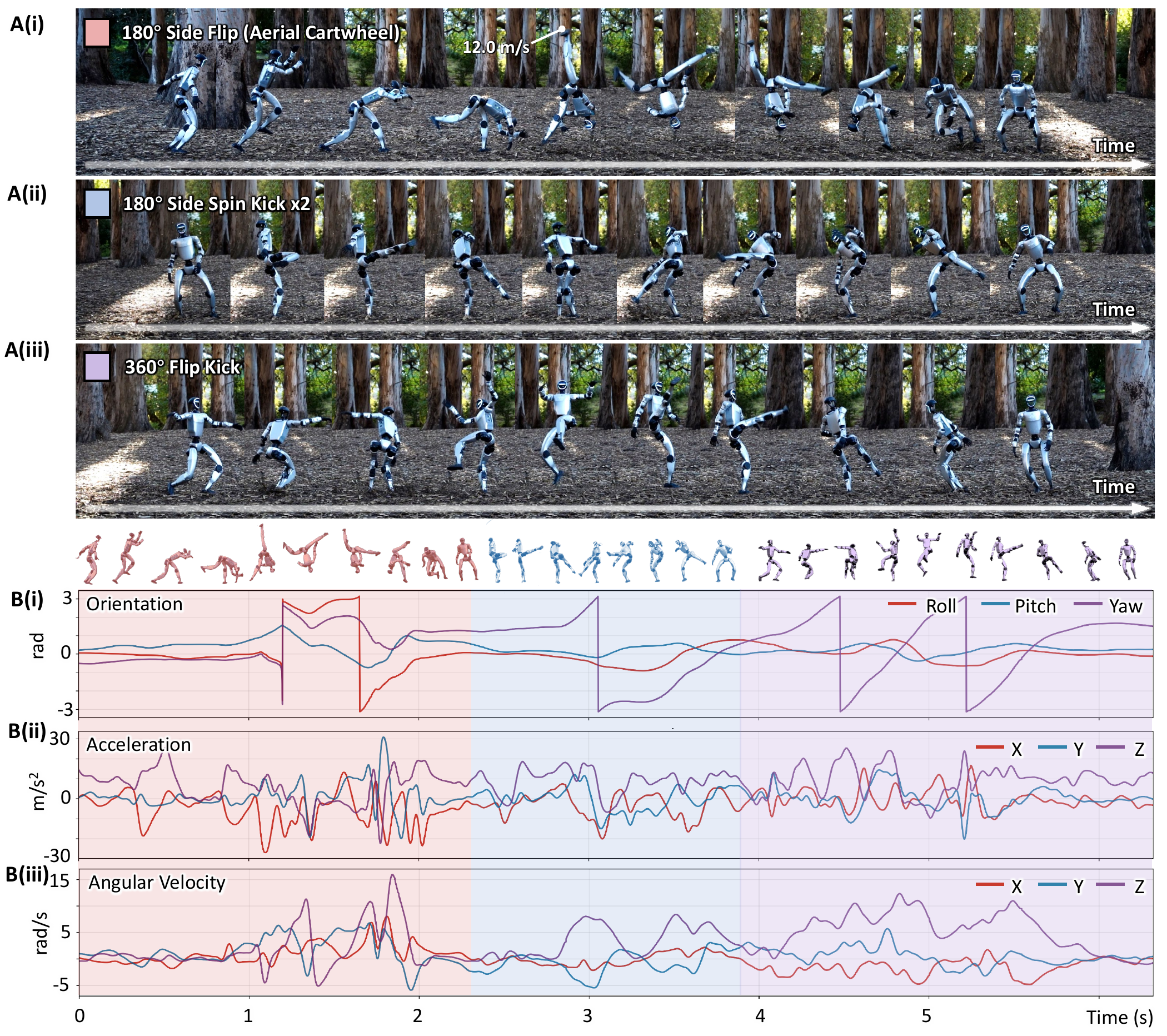}
    \caption{\textbf{Human-level Agility exhibited in a highly dynamic acrobatic motion.} (\textbf{A}) Real-world forest demonstration showing one continuous motion including: \(180^\circ\) side flip (aerial cartwheel), two consecutive \(180^\circ\) side spin kicks, and a \(360^\circ\) flip kick.
    (\textbf{B}) Robot's orientation, linear acceleration, and angular velocity, illustrating highly dynamic behavior; red, blue, and purple shaded intervals denote the \(180^\circ\) side flip, the two \(180^\circ\) side spin kicks, and the \(360^\circ\) flip kick, respectively. 
    }
    \label{fig:agile}
\end{figure}

We evaluated agility in outdoor environments with soft soil, decaying leaves, and uneven ground, introducing deformable and unstable contacts that were not present during training. 
Despite these unseen challenges, the robot completed complex acrobatic sequences and martial-arts-inspired motions (Fig.~\ref{fig:agile}), including aerial cartwheels, a skill that requires explosive takeoff, controlled mid-air rotation, and accurate landing, and is difficult even for trained adult humans.
Specifically, during the airborne phase, the robot reached a peak acceleration of $31\, \mathrm{m/s^2}$ 
and a pelvic angular velocity of up to $20\,\mathrm{rad/s}$ (mean $7.01\,\mathrm{rad/s}$). We noted that comparable values have been reported in skilled human aerial motions ($7.75\,\mathrm{rad/s}$ on average)~\cite{walker2019application}. 
Furthermore, the landings were clean and required minimal recovery, indicating accurate airborne posture control. These results demonstrated that our framework achieved human-level agility in highly dynamic motions.

Our framework further exhibited human-like agility in contact-rich control. This includes two consecutive cartwheels, crawling on the ground, and jumping up from the ground. It further reproduced expressive athletic gestures that require coordinated whole-body timing, such as Cristiano Ronaldo’s celebration jump-turns, and was able to repeat the celebration five times in succession without loss of stability or style, whereas prior work~\cite{he2025asap} reported only a single execution.

\subsubsection*{Natural, Human-like Behaviors}

\begin{figure}
    \centering
    \includegraphics[width=1.0\textwidth]{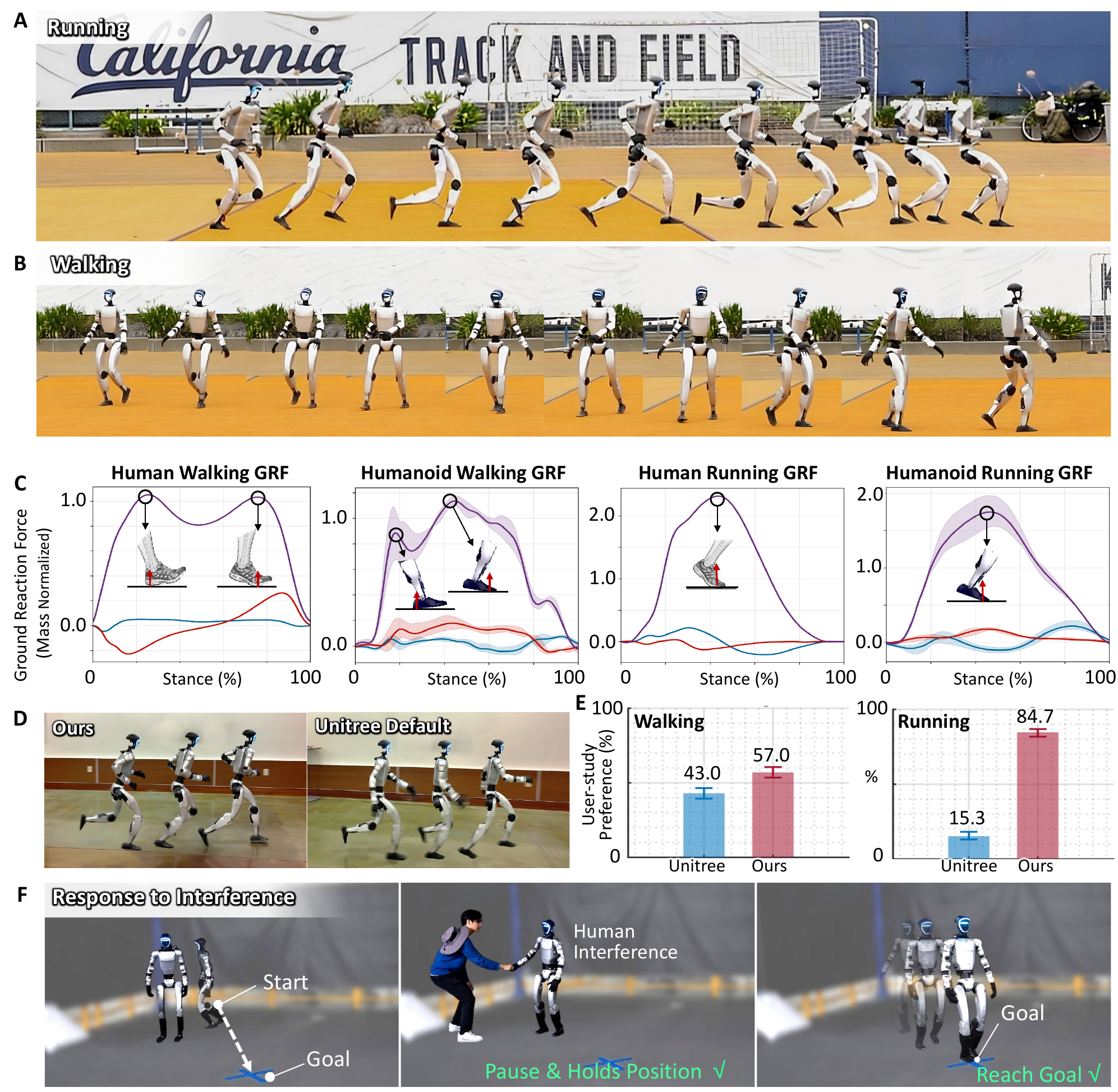}
    \caption{\textbf{Human-like naturalness in walking and running motion.} (\textbf{A}) Natural running on a sports field. (\textbf{B}) Natural walking on a sports field. (\textbf{C}) Comparison of weight-normalized GRF profiles for human vs.\ humanoid during walking and running, showing comparable shapes, contact forces, and timing. (\textbf{D}) Example video clips used in the user study. (\textbf{E}) User-study results for the question ``Which looks more human-like and natural?", comparing our policies vs.\ Unitree’s native walking and running controller, showing that ours is significantly more preferred. The error bars indicate 95\% confidence intervals. (\textbf{F}) Natural response to interference during walking to a goal.}
    \label{fig:natural}
\end{figure}

Our framework demonstrated human-like natural behaviors across diverse tasks such as walking and running, as shown in Fig.~\ref{fig:natural}A-B. To assess biomechanical similarity, we compared Ground Reaction Force (GRF) profiles from the robot and humans using force-sensing treadmills~\cite{Fukuchi2017, horsak2020gaitrec}, normalized by total body weight. As shown in Fig.~\ref{fig:natural}C, the robot exhibited comparable GRF shapes, including double peaks with a distinct heel-strike peak and a propulsive push-off peak in walking, and single peaks in running, with aligned loading and push-off timing. In walking, the robot showed sharper force peaks, reflecting reduced compliance and leg-spring behavior. We attributed this to the lack of a toe joint, which limited rollover and push-off during stance. In running, this limitation was less pronounced because contact occurred only once per step and for a shorter duration, reducing the need for a compliant toe-off phase.

Furthermore, to assess perceptual naturalness quantitatively, we conducted a user study ($N=77$) comparing our motions with Unitree’s native controller, which is the state of the art for this robot.
Participants viewed 20 paired 5-second clips of walking and running from our framework and from Unitree’s native controller, selecting which motion appeared “more human-like and natural.” We conducted a two-tailed binomial test to examine overall preference $(\alpha = .05)$, with separate tests for each gait type using Bonferroni correction $(\alpha = .025)$. BeyondMimic was strongly preferred overall (70.8\% vs. 29.2\%, $p < .001$, Cohen’s $h = 0.859$), with significant preferences in both walking (57.0\% vs. 43.0\%, $p < .001$, $h = 0.281$) and running (84.7\% vs. 15.3\%, $p < .001$, $h = 1.532$).

Finally, we demonstrated human-like recovery under external disturbances (Fig.~\ref{fig:natural}F). When the robot was gently held while walking, it remained compliant, paused and stabilized in place, and then smoothly resumed walking once released, avoiding stiff or exaggerated reactions.

\subsection*{Versatile Humanoid Control}

A key advantage of our framework is its ability to synthesize agile, human-like skills at inference time via diffusion guidance, enabling the completion of unseen tasks without retraining. Figure~\ref{fig:versatile_loco}A illustrates this predictive control process, where the model first predicts future states and actions, then iteratively refines its predictions and converges toward a trajectory that minimizes the specified velocity cost.
Importantly, this approach differs from online trajectory optimization: because the model has already acquired a diverse and feasible set of motor skills as a prior, simple, task-specific costs are sufficient to trigger appropriate behaviors for unseen tasks. This eliminates the need for numerous regularizations and behavior-shaping terms required in either model-based or learning-based methods, providing a much more versatile solution. 
Our experiments demonstrate this versatility in three aspects: command-conditioned locomotion with velocity and waypoint commands, inpainting using keyframes for agile skill composition, and flexible task composition (Fig.~\ref{fig:versatile_loco}, Fig.~\ref{fig:versatile_inpaint_sdf}, and Movie S5, Movie S6).

\subsubsection*{Command-conditioned Locomotion}

\begin{figure}
    \centering
    \includegraphics[width=1.0\textwidth]{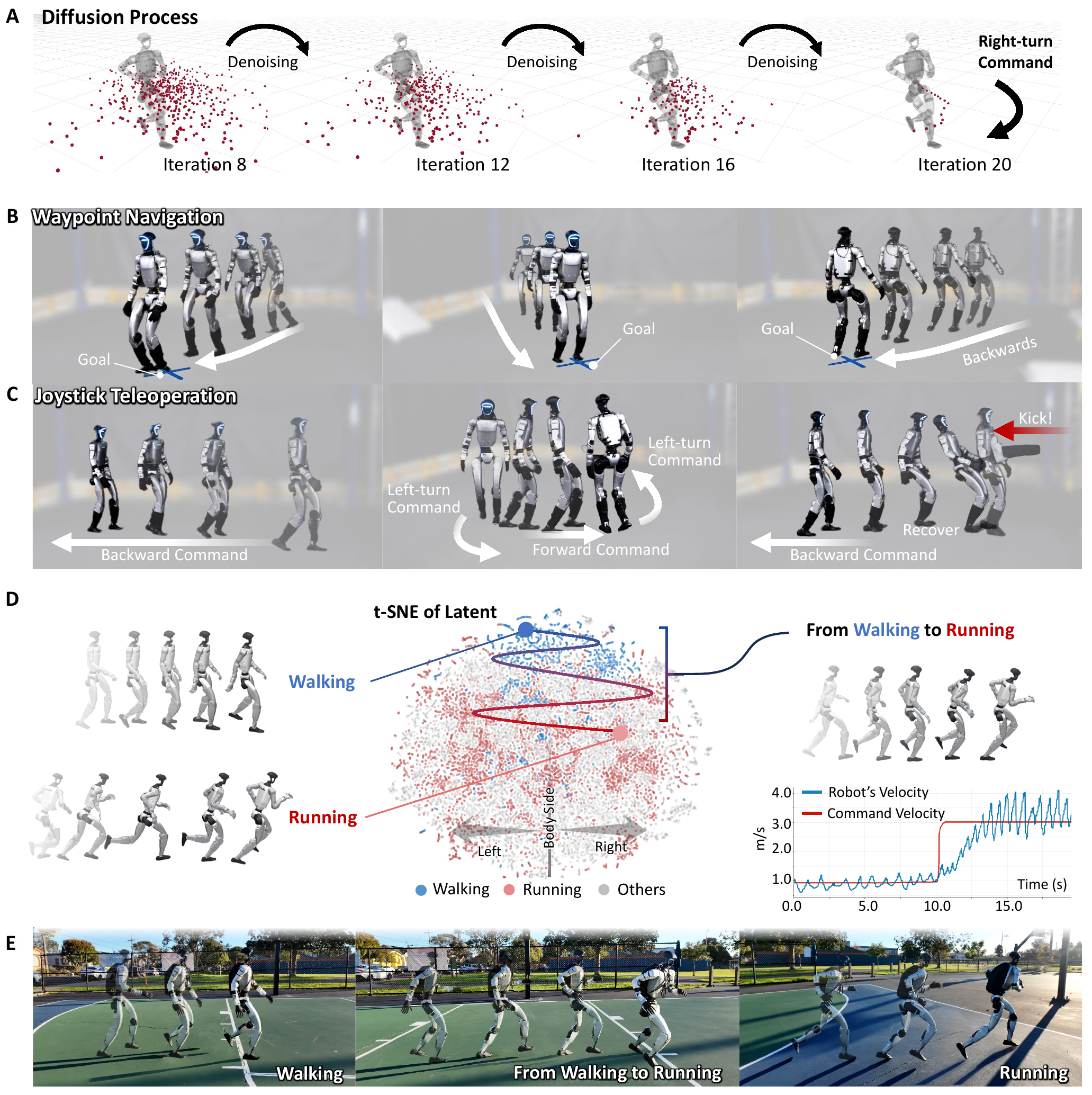}
    \vspace{-1.2cm}
    \caption{\textbf{Command-conditioned locomotion via guided diffusion.}
    (\textbf{A}) Visualization of diffusion process under joystick control. Starting from Gaussian noise, the distribution progressively converges to optimize for a right-turn command.
    (\textbf{B}) Waypoint navigation. From multiple start points, the robot reaches the goal using forward or backward walking.
    (\textbf{C}) Joystick teleoperation. The robot tracks the joystick velocity command. Even under an impulsive disturbance, it recovers quickly and continues following the backward command.
    (\textbf{D}) t-SNE visualization of the latent space, illustrating the transition from walking to running.
    (\textbf{E}) Real-world transition from walking to running conditioned on velocity command.
    }
    \label{fig:versatile_loco}
\end{figure}

First, we demonstrated zero-shot commanded locomotion using either desired velocities or waypoint goals. In this task, the policy received joystick commands specifying linear and yaw velocities, or waypoint targets indicating desired positions (Fig.~\ref{fig:versatile_loco}). Under waypoint navigation (Fig.~\ref{fig:versatile_loco}B), it generated smooth and stable trajectories to reach the goal from varying initial positions. Under joystick control (Fig.~\ref{fig:versatile_loco}C), the robot exhibited smooth omnidirectional walking and reliably tracked the commanded direction. The controller remained robust to large external perturbations, such as kicks, while continuing to pursue the task objectives. For a longer-horizon evaluation, the robot was able to run continuously for over 50 m on a track, indicating stable control over extended distances. We recorded an average velocity tracking error of 12.14\% and 13.65\% in walking and running when evaluated in simulation.  

The policy also demonstrated multimodal locomotion behaviors, producing distinct gaits under similar commands. For example, a low-velocity command led to either a stable walking gait or a light jogging gait, both human-like and dynamically stable. Moreover, given only a desired velocity input, the policy achieved smooth gait transitions from walking to running (Fig.~\ref{fig:versatile_loco}D-E). Notably, such transitions were rare and unlabeled in the motion data; yet our controller inferred and reproduced them naturally when the task required it, much like humans acquiring smooth skill transitions through context and intention rather than explicit skill specification.

\subsubsection*{Motion Inpainting and Task Transitioning}
Having established multimodal locomotion under velocity commands and waypoints, many agile acrobatics demand stronger, temporally structured objectives. 
We therefore move beyond weak velocity or position cues and instead use motion inpainting, where a sparse set of future keyframes guides the policy to generate smooth intermediate motions, enabling online insertion, transition, and composition of agile motor skills.
As illustrated in Fig.~\ref{fig:versatile_inpaint_sdf}A, starting from joystick-controlled walking, we injected desired cartwheel keyframes at 0.2 s intervals.
Consequently, the diffusion policy drove a smooth transition into the cartwheel and inpainted the discrete keyframes into a temporally coherent and continuous trajectory. Upon completion of the cartwheel, the controller returned to command conditioned walking smoothly. 

Following the same inpainting setup, the diffusion model was able to accurately demonstrate a diverse range of challenging motions learned from human motion tracking, along with the emergent transitions into these motions. Starting from command-conditioned locomotion, the model transitioned smoothly into these complex skills, including contact-rich behaviors such as walking into lying down and agilely getting back up to standing, as well as highly dynamic gaits such as spin-kicks and flip-kicks.  

Beyond the versatility to compose agile motions, our unique online optimization–based framework further enabled the versatility to transition across different task specifications. we demonstrated three consecutive cartwheels stitched with walking and running before and after, as shown in Fig.~6A(iii). This long-horizon execution showcased not only the robustness and smoothness in mode switching, but more importantly the capability to switch back and forth freely between different tasks such as velocity following and motion inpainting. 

\subsubsection*{Task Composition}
Beyond transitioning, our framework demonstrated intuitive task composition, a challenge for most prior goal-conditioned controllers. This challenge arises mainly from the combinatorial growth of possible goal combinations to enumerate. Moreover, certain tasks are difficult to be fully covered during training but can be efficiently evaluated at test time. Our approach addressed this by allowing multiple objectives to be evaluated and optimized at inference time without enumerating all possibilities. Because the costs remained simple and task-specific, they could be flexibly summed up without creating a complex objective that requires extensive tuning, enabling the framework to solve compositions of unseen tasks without retraining.

As an example of versatility in task composition, we achieved simple scene-aware navigation by combining a waypoint-tracking cost with an obstacle-avoidance cost, as shown in Fig.~\ref{fig:versatile_inpaint_sdf}B. The obstacle-avoidance cost, shaped by a signed distance field (SDF), provides distance gradients across the prediction horizon at each denoising step, steering the predicted trajectory away from obstacles. As a result, the robot successfully detoured around the obstacle and reached the target waypoint. This same cost composition remained effective when a joystick-tracking cost replaced the waypoint term, enabling the system to mitigate collisions even under moderate off-goal user inputs.


\section*{Discussion}
The presented results substantially advance the published state of the art in humanoid control. 
BeyondMimic is a scalable
pipeline that learns directly from human motion to reproduce agile, natural behaviors and synthesize them into task-directed actions -- all without task-specific tuning. By learning from diverse human motion data, our model acquires hundreds of skills that demonstrate human-level agility while maintaining stylistic and naturalistic characteristics. Moreover, the model does not merely memorize complete motion sequences; it composes atomic behaviors seamlessly to produce novel trajectories.

Before our work, a prevailing assumption in humanoid control was that achieving robust real-world transfer required heavy domain randomization, manual gain tuning, and carefully engineered regularization to counteract the large sim-to-real gap. These efforts often involved random torque perturbations, simulated delays, or delicately tuned reward shaping terms penalizing contact forces, slippage, and stumbles. In contrast, we show that such extensive heuristics are unnecessary. With a principled reward formulation, moderate domain randomization, and careful system implementation, our framework learns hundreds of diverse motions under a single RL formulation and one shared set of hyperparameters. The same policy transfers directly to hardware—without motion- or robot-specific tuning—while retaining both agility and natural appearance.

More importantly, BeyondMimic moves beyond imitation. For the first time, we demonstrate that a unified model learned from unlabeled human motion can synthesize and compose its skills into coherent sequences that directly solve novel downstream tasks. 
Through guidance in the denoising process, the controller predicts and adjusts future trajectories to optimize for new objectives at inference time, effectively coupling planning and control within a single model. 
As a result, the same model that performs sprinting or cartwheeling can autonomously navigate around obstacles, follow joystick-driven velocity commands, or pause and recover under external disturbances, while preserving smooth, human-like behaviors.

The training process is fully task-agnostic and requires no labeled data, providing a scalable path toward general-purpose humanoid control. This design enables continuous extension: new motion data expands the skill set without requiring redefinition of objectives or retraining for each task. The open-source release of BeyondMimic has been widely adopted, serving as a default method in public reinforcement learning repositories such as MJLab and Unitree RL Lab\cite{zakka2025mjLab, unitree2025untreerllab}. More broadly, this work expands the domain of humanoid control from manually tuned, motion-specific policies to a predictive, generalizable paradigm that unifies high-quality human-like behaviors with control and planning. We view this as a step toward foundational models for humanoid behavior, models that learn directly from human demonstrations and generalize across motions, environments, and objectives.

\subsubsection*{Limtations and Future Works}
Our approach has several limitations that present opportunities for future work. The diffusion model inherits the quality of the underlying state estimation system, such that errors in proprioception directly propagate to generated trajectories. While our latent diffusion formulation provides some robustness to noisy observations, improving state estimation through sensor fusion or learned estimators remains a promising direction for enhanced performance.

Beyond perception noise, the prediction capability could also be improved. The current system predicts trajectories over a 0.64-second horizon, which is sufficient for reactive control and local obstacle avoidance but insufficient for long-horizon planning tasks that require reasoning about distant goals or obstacles requiring early anticipation. 
In addition, the inclusion of history is critical for stabilizing future predictions but can cause the model to become trapped in repetitive motion patterns
even when guidance is active. To counteract this, we apply larger guidance weights, which can, however, destabilize the denoising process during mode switching or in high-variance states. As a result, under guided diffusion, 
the robot is stable once a gait orbit is established, but tends to stumble at the start and end of motions. Future work could explore removing or reducing this dependence on history to improve responsiveness and transient behaviors.

Finally, at the task level, our guidance-based optimization works well for coarse-grained objectives but less so with fine-grained ones, and still requires light-weight tuning of the guidance weights. Future work could explore established approaches for fine-grained control over diffusion models, such as supervised fine-tuning~\cite{ruiz2023dreambooth, brooks2023instructpix2pix} and adapter-style control layers~\cite{hertzprompt, zhang2023adding, mou2024t2i}, which have proven highly effective in vision domains and could potentially enable fine-grained, composable trajectory control without extensive retraining or manual weight tuning.

\section*{Materials and Methods}

\subsection*{Overview}
The objective of our model is to achieve versatile humanoid control on various unseen downstream tasks, synthesizing diverse motions with sustained agility and human-like naturalness.

In the first stage, we focus on learning a diverse set of human motions through scalable motion tracking with RL. Prior works either trained a single multi-skill policy~\cite{he2025omnih2o, ze2025twist}, which is scalable but produces unnatural behaviors due to insufficient RL exploration, or learned separate motion-specific policies~\cite{xie2025kungfubot, zhang2025hub} that achieve natural motions but require motion-specific tuning. Contrary to these methods, we find that a general, simple yet efficient RL pipeline is sufficient for scalable motion tracking. With a shared set of hyperparameters, the pipeline preserves human-level agility and naturalness while providing the scalability and consistency needed to extend across diverse motions.

In the second stage, we train a unified diffusion model that integrates diverse motions to enable expressive skill composition and online task optimization in a predictive control manner. The key motivation for using diffusion models is that they naturally support predictive control through classifier guidance, an online optimization process that steers unconditional generation toward task-specific conditional generation. Unlike prior diffusion-based action-only policies~\cite{Wang2025HDC, xiayou_Diffuseloco}, the key to enabling this capability is the adoption of a latent state–action diffusion model, which implicitly captures how actions influence future states and provides foresight during inference. 
This predictive structure enables versatile control in unseen scenarios by synthesizing learned skills seamlessly.

Trainings are conducted entirely in simulation using the most accurate manufacturer-specified parameters, without any additional system identification. After training, each stage is deployed zero-shot on physical robots. A key enabler of this transfer is our deployment framework, developed entirely in C++ and optimized for real-time execution. This framework ensures stable, low-latency control and precise synchronization between policy inference and hardware execution---addressing a critical source of the sim-to-real gap. Despite running only on CPUs and a modest mobile GPU, this implementation enables strong sim-to-real transfer for both stages without any motion- or robot-specific tuning.

\subsection*{Scalable Human Motion Tracking via RL}
We define scalable motion tracking as a one-recipe-fits-all framework, where each motion is trained with its own policy but under a shared formulation and training setup, minimizing the tuning effort for any new motion without compromising motion quality.
To achieve this goal, we present a principled, motion-agnostic pipeline for tracking human motions with human-like agility and naturalness. We formulate the motion-tracking problem as a Markov Decision Process (MDP) and solve it using RL, which maximizes the expected cumulative reward under this MDP. Given minutes-long reference motions, the pipeline produces sim-to-real-ready motion-tracking policies using the same MDP and hyperparameters across all motions. This unified setup enables seamless scaling to hundreds of skills without manual tuning.

Since our MDP formulation is time-invariant, we omit the timestep \(t\) in the following formulation for clarity. The subscript `\(\text{ref}\)' denotes quantities from the reference motion. Unless otherwise specified, all quantities in this section are expressed in the world frame.

\subsubsection*{Tracking Objective}

Given a human-retargeted reference motion, the goal is to reproduce it on a real humanoid robot with high fidelity, while tolerating global drift caused by perturbations and sim-to-real mismatch.
We start from the motion data $(\mathbf{q}^{\text{ref}}, \boldsymbol{\nu}^{\text{ref}})$, represented as per-frame generalized positions $\mathbf{q}^{\text{ref}}=(\mathbf{p}^{\text{ref}}, R^{\text{ref}}, \boldsymbol{\theta}^{\text{ref}})\in \mathbb{R}^3\times \operatorname{SO}(3)\times \mathbb{R}^{n_{\text{jnt}}}$ and velocities $\boldsymbol{\nu}^{\text{ref}}=(\mathbf{v}^{\text{ref}}, \boldsymbol{\omega}^{\text{ref}}, \dot{\boldsymbol{\theta}}^{\text{ref}})\in \mathbb{R}^3\times \mathbb{R}^3\times \mathbb{R}^{n_{\text{jnt}}}$, where $n_{\text{jnt}}$ is the number of joints for the robot. Forward kinematics yields the pose \(T_b^{\text{ref}}=(\mathbf p_b^{\text{ref}}, R_b^{\text{ref}})\) and the twist \(\mathcal V_b^{\text{ref}}=(\mathbf v_b^{\text{ref}}, \boldsymbol\omega_b^{\text{ref}})\) for each body link \(b \in \mathcal B\), where $\mathcal{B}$ is the set of all robot bodies. 
To avoid redundancy from closely spaced links, a compact set of target bodies \(\mathcal B_{\text{target}} \subset \mathcal B \), including the end-effectors $\mathcal{B}_\text{ee} \subseteq \mathcal B_{\text{target}} $, is selected for tracking.

Perturbations for robustness during training and the sim-to-real gap inevitably lead to global drift. To preserve motion style while allowing such drift, the policy should track relative body poses instead of global ones.
As shown in Fig.~\ref{fig:pipline}A, we define an anchor body $b_\text{anchor}$ (typically the root or torso) and use it to express desired poses of the bodies in the \emph{anchor-centered} frame.
The anchor itself follows the reference directly, $T_{\text{anchor}}^{\text{des}} = T_{\text{anchor}}^{\text{ref}}$.
For any non-anchor body \(b\neq b_{\text{anchor}}\), we set $T_b^{\text{des}} = \mathcal A\!\left(T_b^{\text{ref}},\, T_{\text{anchor}}\right)$, where \(\mathcal A(\cdot)\) is a yaw-aligned, height-preserving transform (see Supp. S1), 
while the desired twists remain unchanged $\mathcal V_b^{\text{des}} = \mathcal V_b^{\text{ref}}$.
The resulting motion-tracking objective is then: 
\[
\bigl(
T_{\text{anchor}}^{\text{des}},\mathcal V_\text{anchor}^{\text{des}},\;
\{\,T_b^{\text{des}},\, \mathcal V_b^{\text{des}}\,\}_{\,b\in\mathcal B_{\text{target}}}
\bigr).
\] 
This objective preserves motion style while allowing benign global drift, improving robustness and sim-to-real transfer.

\subsubsection*{Rewards}
To maximize transferability across motions and minimize motion-specific bias, we design a simple, motion-agnostic reward composed of a unified task term and three regularization penalties. 
The task rewards adopt a unified, task-space formulation with uniform weighting to promote accurate body tracking across all target bodies.

We compute tracking errors for position, orientation, linear velocity, and angular velocity, $\mathbf{e}_{s}$ for $s \in \{\mathbf{p}, R, \mathbf{v}, \boldsymbol{\omega}\}$ between the desired and actual poses and twists. For each $s$, we take the mean-squared error $\bar{e}_{s}$, over all target bodies $b \in \mathcal{B}_\text{target}$.
Each $\bar{e}_{s}$ is mapped through a Gaussian-shaped exponential: $r(\bar{e}_{s},\sigma_s) = \exp\left(-{\bar{e}_{s}}/{\sigma^2_s}\right)$, with nominal error $\sigma_s$ for each term determined empirically. 
The total task reward is defined as
\[
r_{\text{task}}
= \sum_{s \in \{\mathbf{p}, R, \mathbf{v}, \boldsymbol{\omega}\}} r(\bar{e}_s, \sigma_s).
\]
Optionally, global tracking terms can be included for the anchor body, following the same formulation as $r_{\text{task}}$ but using only the position and orientation errors, ${\mathbf{e}}_{\mathbf{p},\text{anchor}}$ and ${\mathbf{e}}_{R,\text{anchor}}$.

In contrast to prior work that uses many ad-hoc regularizations, we use only three lightweight, broadly applicable penalties that generalize across motions to encourage physically consistent behaviors. First, the joint limit penalty, $r_{\text{limit}}$, encourages joint positions to remain within soft limits to avoid damaging the hardware. 
Second, the action rate penalty, $r_{\text{smooth}}$, promotes smooth transitions between consecutive actions, preventing policies with excessive jitter. 
Third, to penalize self-collisions, we count the number of bodies whose self-contact forces exceed a predefined threshold and sum them as the total contact penalty, $r_{\text{contact}}$, over all bodies $b \notin \mathcal{B}_\text{ee}$. 
The total reward is then defined as
\[
r = r_{\text{task}} - \lambda_{l} r_{\text{limit}} - \lambda_{s} r_{\text{smooth}} - \lambda_{c} r_{\text{contact}},
\]
where $\lambda_{l}, \lambda_{s}, \lambda_{c} > 0$ denote the corresponding weights. The weights and hyperparameters are detailed in S1. 

\subsubsection*{Observation and Action}
We use a continuous, robot-centric observation space without temporal stacking, which simplifies training and improves sim-to-real transfer.
The observation for policy input is
\[
\mathbf{o}=[\boldsymbol{\psi},\,\mathbf{e}_{\text{anchor}},\,\mathcal{V}_{\text{imu}},\,\boldsymbol{\theta}- \boldsymbol{\theta}^{0},\,\dot{\boldsymbol{\theta}},\,\mathbf{a}_{\text{last}}],
\]
which consists of: 
\textbf{(i) Motion phase} from the reference motion,
\(\boldsymbol{\psi}=[\boldsymbol{\theta}^{\text{ref}},\,\dot{\boldsymbol{\theta}}^{\text{ref}}]\), 
used solely as a progress cue; the policy is not intended to track these joint states directly. 
\textbf{(ii) Anchor pose error} 
\(\mathbf{e}_{\text{anchor}}\in\mathbb{R}^{9}\), 
stacks the position error \({\mathbf{e}}_{\mathbf{p},\text{anchor}}\),
a 6-D orientation error obtained by taking the first two columns of the rotation error matrix \(R_{\text{anchor}}^{\text{des}}R_{\text{anchor}}^{\top}\) (Rot6D ~\cite{zhou2019continuity}); since the reference is defined globally, this term provides the minimal global cue needed for balance and drift correction. 
\textbf{(iii) Other proprioception} includes the IMU twist expressed in the IMU frame \(\mathcal{V}_{\text{imu}}\in\mathbb{R}^{6}\), which aid push recovery, foot timing, and stabilization; joint states including joint positions relative to the default \(\boldsymbol{\theta}-\boldsymbol{\theta}^{0}\) and joint velocities \(\dot{\boldsymbol{\theta}}\); and the previous action \(\mathbf{a}_{\text{last}}\). The joint state paired with the previous action offers a proxy for the joint torque and contact currently being realized. In addition, the previous action with the action-smoothness penalty \(r_{\text{smooth}}\) helps suppress high-frequency jitter. 

Action is designed as normalized joint position setpoints: $\boldsymbol{\theta}^{\text{sp}} = \boldsymbol{\theta}^{0} + \boldsymbol{\alpha}\odot \mathbf{a}$, where $\mathbf{a}\in\mathbb{R}^{n_\text{jnt}}$ is the policy output and $\boldsymbol{\alpha}$ is a per-joint action scale ($\odot$ represents elementwise product). 
These setpoints are sent to the low-level motor drivers as position PD controller commands to generate torques.
However, they are not position targets or plans to be tracked with high precision; rather, they act as intermediate variables shaping the desired torques and are intentionally not clipped by joint kinematic limits.

High joint impedance (PD gains) is common in prior work to simplify control~\cite{jason_deepMimic,PHC,he2025asap, ze2025twist}: in free space, the high feedback gains suppressed the natural dynamics and dominate the closed-loop dynamics, causing the joints to behave like stiff, high-bandwidth position servos—i.e., tracking becomes close to kinematic playback. 
However, such high-impedance settings are typically impractical on hardware: they amplify sensor noise, decrease passive compliance needed for impact absorption, and obscure implicit torque information carried by current joint state and prior commands.
In Supplementary S1, we detail a heuristic procedure for selecting reasonable joint impedances and action scales on high-DoF humanoid systems; In Supplementary S2, we include ablations on joint impedance tuning.

\subsubsection*{Domain Randomization}
Domain randomization is essential for successful sim-to-real transfer. However, excessive randomization can degrade policy performance, making behaviors overly conservative and training unnecessarily difficult. To enable scalable training across diverse motions while preserving smooth and natural behavior, we find that maintaining a compact set of domain randomizations is most effective. In practice, we randomize three key parameters that generalize well across motions: the ground friction and restitution coefficients, the default joint positions $\boldsymbol{\theta}^0$ (for both actions and observations, simulating joint offset calibration errors), and the torso center of mass position. Additionally, random velocity perturbations are applied during training to improve robustness to environmental variability. The details for domain randomizations are in S1. 

\subsubsection*{Adaptive Sampling}
Training long motion sequences presents the challenge that different segments vary widely in difficulty. 
Uniformly sampling across the entire trajectory, as commonly done in prior works~\cite{he2025asap, PHC}, often oversamples easy portions while undersampling harder ones, leading to slow learning or even failure to converge. 
To address this, we employ an adaptive sampling strategy that prioritizes segments with higher empirical failure rates. 
After the policy masters the difficult segments and the failure rates decrease, the sampler gradually reverts to uniform sampling to preserve coverage of easier regions.

\subsection*{Versatile Humanoid Control via Guided Diffusion}
The key to achieving versatile control is enabling online optimization for tasks unseen during training. Diffusion models~\cite{ho2020denoising}, beyond capturing multimodal distributions, can optimize their outputs toward novel objectives via gradient ascent, a process known as classifier guidance. First introduced for image generation\cite{nichol2021glide}, it has been proven effective in RL~\cite{janner2022planning} and character animation~\cite{karunratanakul2023guided,cohan_flexible_motion_inbetween}. However, using this capability is nontrivial, as task objectives are defined in the state space, whereas typical policies output in the action space. One approach is to use forward dynamics models~\cite{xue2025full, roth2025learned} to bridge this gap, but such models are often constrained by high dimensionality and real-time requirements. Inspired by prior works, we instead bridge this gap by jointly modeling states and actions in a latent diffusion model with causal consistency between them. Moreover, instead of modeling only the next step, the model predicts a short horizon of future trajectories, allowing task objectives to steer the future states, which in turn produce the corresponding actions in a receding-horizon control fashion.

Shown in Fig.~\ref{fig:pipline}B, the latent state–action diffusion model consists of two parts. First, we train a Variational Autoencoder (VAE) using a diverse set of motion-tracking policies, which provides smooth motion representations for the diffusion model. Second, we train a state–latent diffusion model using latents collected by rolling out the VAE. Both components are trained on unlabeled, task-agnostic data. At inference time, we apply an unseen cost function to steer the state–latent diffusion model toward desired trajectories and decode its latent outputs with the VAE decoder to produce joint-level actions.

\subsubsection*{Latent Diffusion Models}

Diffusion models are a class of generative models that generate samples by reversing a gradual noising process~\cite{ho2020denoising}. 
In the forward process, a clean sample $\mathbf{x}_0$ is progressively corrupted by Gaussian noise, and in the reverse process, pure Gaussian noise is iteratively denoised using Stochastic Langevin Dynamics~\cite{welling2011bayesian} to produce a clean sample. 

Building upon this framework, Latent Diffusion Models (LDMs)~\cite{rombach2022high} perform diffusion in a lower-dimensional latent space, typically obtained via a VAE with encoder $\mathcal{E}$ and decoder $\mathcal{D}$ that map between data and latent spaces as $\mathbf{z} = \mathcal{E}(\mathbf{x})$ and $\hat{\mathbf{x}} = \mathcal{D}(\mathbf{z})$. The VAE is trained by maximizing the evidence lower bound (ELBO):
\[
\mathcal{L}_{\text{VAE}} = \mathbb{E}_{q_{_{\mathcal{E}}}(\mathbf{z}|\mathbf{x})}\left[\|\mathbf{x} - \mathcal{D}(\mathbf{z})\|^2\right] 
+ \beta\, D_{\mathrm{KL}}\left(q_{_{\mathcal{E}}}(\mathbf{z}|\mathbf{x}) \,\|\, \mathcal{N}(\mathbf{0}, \mathbf{I})\right),
\]
where $D_{\mathrm{KL}}$ denotes the Kullback--Leibler divergence~\cite{kullback1951information}, and $q_{_{\mathcal{E}}}(\mathbf{z}|\mathbf{x})$ is the encoder’s posterior distribution. The first term promotes accurate reconstruction, and the second, scaled by $\beta$, regularizes the latent space. 

The LDM used in this work follows the denoising diffusion probabilistic model (DDPM)~\cite{ho2020denoising}, which defines a noise schedule parameterized by coefficients $\{\alpha_k, \gamma_k, \sigma_k\}_{k=1}^K$, controlling the signal retention and noise magnitude up to the maximum diffusion step $K$. Let $\mathbf{z}^k$ denote the noisy latent at diffusion step $k$. In the forward diffusion process, the clean data $\mathbf{x}^0$ is first encoded into its latent representation $\mathbf{z}^0 = \mathcal{E}(\mathbf{x}^0)$, and then noised to step $k$ following the diffusion posterior distribution,
\[
q_\text{forward}(\mathbf{z}^k \,|\, \mathbf{z}^0) =
\mathcal{N}\!\left(\sqrt{\bar{\alpha}_k}\,\mathbf{z}^0,\; (1-\bar{\alpha}_k)\mathbf{I}\right),
\]
where $\bar{\alpha}_k = \prod_{i=1}^{k}\alpha_i$. 
Then, a neural network $z_{\phi}(\mathbf{z}^k, k)$ 
is trained to predict the clean latent from the noisy input via
\[
\mathcal{L}_\text{Diffusion} = \mathbb{E}\!\left[\|z_{\phi}(\mathbf{z}^k, k) - \mathbf{z}^0\|^2\right].
\]
During inference, the reverse process iteratively denoises Gaussian noise back into a clean latent:
\[
\mathbf{z}^{k-1} =
\alpha_k\!\left(\mathbf{z}^k - \gamma_k\big(\mathbf{z}^k - z_{\phi}(\mathbf{z}^k, k)\big)\right)
+ \sigma_k\,\mathcal{N}(0, \mathbf{I}).
\]
Finally, the decoder $\mathcal{D}$ maps the recovered latent $\mathbf{z}_0$ back to the data space as $\hat{\mathbf{x}}^0 = \mathcal{D}(\mathbf{z}^0)$.

\subsubsection*{LDMs for Trajectory Modeling}
In our setting, predictive control with human-like behavior amounts to learning the distribution of trajectories that give rise to coordinated, human-like motion. We formulate this as a trajectory modeling problem and use an LDM to represent it. The trajectories are generated by motion-tracking policies and serve as samples of general human-like behavior. Importantly, they include only state-action pairs, excluding any reference information, since the goal is to model the intrinsic distribution of human-like behaviors rather than the tracking process itself.

To ensure spatial consistency, we use a hybrid character–yaw-centric parameterization for the state space. A trajectory consists of $N$ past timesteps, current timestep, and $H$ future timesteps. 
For each timestep \( n \in [-N, H] \) in the trajectory, the state \( \mathbf{s}_{t+n} \) 
contains root pose $(\mathbf{p}_{\text{imu}}^{\,t+n}, \mathbf{R}_{\text{imu}}^{\,t+n})$ and twist $(\mathbf{v}_{\text{imu}}^{\,t+n}, \boldsymbol{\omega}_{\text{imu}}^{\,t+n})$ 
expressed with respect to the current root frame at timestep \( t \), 
and body positions \( \mathbf{p}_{\text{b}}^{\,t+n} \) and velocities 
\( \mathbf{v}_{\text{b}}^{\,t+n} \) for bodies \( b \in \mathcal{B}_\text{target} \) expressed relative to their local root frame at timestep \( t{+}n \). 
The action space is the same as in motion-tracking policies, which correspond to the target inputs of the impedance controller. Details for state parameterization are provided in Section~S2.

Given the trajectory modeling problem, the motivations to use an LDM over a standard diffusion model are twofold. First, the action space, serving as setpoints for the impedance controller, is highly irregular with sharp torque spikes, making direct diffusion learning unstable and violating the smoothness assumption of diffusion models. Second, large diffusion networks required for accurate trajectory modeling introduce non-negligible inference latency, causing generated actions to lag behind the latest states. In contrast, a latent space with smooth motion representations is well suited for diffusion modeling, and a lightweight decoder can leverage up-to-date observations to transform them into precise, dynamically consistent actions.

Learning an LDM from human motions consists of two stages. 
First, we acquire motion representations in a latent space by imitating motion tracking policies that receive observation $\mathbf{o}$ and output action $\mathbf{a}$. 
Inspired by prior work~\cite{luouniversal}, we encode reference motions with a conditional VAE instead of raw PD actions with a regular VAE, 
since motion states are more structured and easier to encode than raw actions.
The encoder receives only the reference-motion components
to produce a latent representation 
\( \mathbf{z} = \mathcal{E}(\boldsymbol{\psi},\, \mathbf{e}_{\text{anchor}}) \) 
that captures motion intents. 
The decoder then combines this latent with other proprioceptive inputs 
to reconstruct the action,  
\[
\hat{\mathbf{a}} = \mathcal{D}(\mathbf{z},\, [\mathbf{g},\, \mathcal{V}_{\text{imu}},\, \boldsymbol{\theta},\, \dot{\boldsymbol{\theta}},\, \mathbf{a}_{\text{last}}]),
\]
where \( \mathbf{g} \) denotes the projected gravity vector expressed in the root frame. 
Subsequently, the VAE is trained with DAgger~\cite{ross2011reduction} using the modified ELBO:
\[
\mathcal{L}_\text{VAE}
= \mathbb{E}_{q_{_{\mathcal{E}}}(\mathbf{z} \mid [\mathbf{c},\, \mathbf{e}_{\text{anchor}}])}\!\left[\|\hat{\mathbf{a}} - \mathbf{a}\|^2\right]
+ \beta\, D_\mathrm{KL}\!\big(q_{_{\mathcal{E}}}(\mathbf{z} \mid \boldsymbol{\psi},\, \mathbf{e}_{\text{anchor}}) \,\|\, \mathcal{N}(\mathbf{0},\mathbf{I})\big).
\]

Second, we roll out the trained VAE on human motions to collect trajectories and encode each action into a latent, yielding state–latent trajectories 
\( \tau = [\,\mathbf{s}_{t-N},\, \mathbf{z}_{t-N},\, \dots,\, \mathbf{s}_{t}\,, \mathbf{z}_{t}\,,\dots,\, \mathbf{s}_{t+H},\, \mathbf{z}_{t+H}\,] \) 
for LDM training. 
We adapt the earlier LDM formulation to this state–latent trajectory and use \textit{individual denoising steps} for each state and latent, 
\(
\mathbf{k} = 
\big[\, 
k_{\mathbf{s}_{t-N}},\, k_{\mathbf{z}_{t-N}},\,
\dots,\,
k_{\mathbf{s}_t},\, k_{\mathbf{z}_t},\,
\dots,\,
k_{\mathbf{s}_{t+H}},\, k_{\mathbf{z}_{t+H}}
\,\big],
\)
enabling varying noise levels across the horizon for inpainting observations and future poses. 
Training for the LDM is self-supervised. 
Given a clean trajectory \( \tau\) 
and uniformly sampled denoising steps 
\( \textbf{k}_{s_i},\, \textbf{k}_{z_i} \sim \mathcal{U}(0, K) \), 
the denoising network \( z_{\phi}(\tau^{\mathbf{k}},\, \mathbf{k}) \) 
is trained to reconstruct the clean trajectory 
from its noised version \( \tau^{\mathbf{k}} \) by minimizing
\[
\mathcal{L}_\text{Diffusion} = 
\mathbb{E}\!\left[\|z_{\phi}(\tau^{\mathbf{k}},\, \mathbf{k}) - \tau\|^2\right].
\]
At inference, trajectories are generated by starting from Gaussian noise and iteratively denoised using
\[
\tau^{\mathbf{k}-1} =
\alpha_{\mathbf{k}}\!\left(\tau^{\mathbf{k}} -
\gamma_{\mathbf{k}}\big(\tau^{\mathbf{k}} -
z_{\phi}(\tau^{\mathbf{k}},\, \mathbf{k})\big)\right)
+ \sigma_{\mathbf{k}}\,\mathcal{N}(0,\,\mathbf{I}),
\]
which progressively denoises the sample into a clean, temporally consistent state–latent trajectory. 
Finally, the current action is decoded from the current denoised latent $\mathbf{z}_t$ using the most up-to-date observations. The detailed architectures and hyperparameters for the LDM are provided in Section S3.

\subsubsection*{Online Optimization via Guidance}

To adapt the task-agnostic LDM trained above to specific tasks at inference time, we perform online optimization via \emph{classifier guidance}. 
Unlike other generative models such as GANs and VAEs that learn an explicit data distribution, diffusion models learn the \emph{gradient field} of the data distribution itself---known as the \emph{score function} $\nabla_{\boldsymbol{\tau}} \log p(\boldsymbol{\tau})$. Building on this property, we can transform the unconditional score function into a \emph{conditional} one by applying Bayes’ rule:
\[
\nabla_{\boldsymbol{\tau}} \log p(\boldsymbol{\tau} \mid \boldsymbol{\tau}^*) 
= \nabla_{\boldsymbol{\tau}} \log p(\boldsymbol{\tau}) 
+ \nabla_{\boldsymbol{\tau}} \log p(\boldsymbol{\tau}^* \mid \boldsymbol{\tau}),
\]
where $\boldsymbol{\tau}^*$ denotes the desired or optimal trajectory. 
This formulation enables gradient-based optimization entirely at inference time---without retraining---provided we can express the conditional gradient term \(\nabla_{\boldsymbol{\tau}} \log p(\boldsymbol{\tau}^* \mid \boldsymbol{\tau})\) in a computable form. 
To achieve this, we approximate the conditional likelihood using a differentiable, task-specific cost function $G(\boldsymbol{\tau})$, which quantifies how well a sampled trajectory satisfies the task objective. 
By associating the likelihood with this cost as \(p(\boldsymbol{\tau}^* \mid \boldsymbol{\tau}) \propto \exp(-G(\boldsymbol{\tau}))\), the conditional gradient simplifies to
\[
\nabla_{\boldsymbol{\tau}} \log p(\boldsymbol{\tau}^* \mid \boldsymbol{\tau})
= -\,\nabla_{\boldsymbol{\tau}} G(\boldsymbol{\tau}).
\]
This practical formulation allows the diffusion process to incorporate arbitrary differentiable cost functions as optimization objectives, enabling versatile control over diverse and unseen tasks---such as joystick control, obstacle avoidance, and motion inpainting---in an inference-time, training-free manner. Specific cost functions for each task are detailed in S3.

\subsection*{Validation of the method}
We present ablation studies to justify key design decisions. For motion tracking, we justify (i) notable MDP parameters, and (ii) adaptive sampling in resets. 

\subsubsection*{MDP Design and Parameters}
We systematically evaluate how different MDP design choices affect sim-to-real performance. To make the evaluation sufficiently challenging and reveal meaningful differences, we select the martial arts motion shown in Fig.~\ref{fig:ablation}A(ii–iii), repeat each experiment three times, and record ground-truth trajectories using a Motion Capture system. The resulting local and global tracking errors are shown in Fig.~\ref{fig:ablation}A.

We first examine the effect of rotation representation in the observation. Baselines using quaternions show higher tracking errors, and the one with axis–angle fails once. In contrast, smooth and continuous forms such as Rot6D achieve markedly better sim-to-real performance. While prior work~\cite{zhou2019continuity} demonstrates the benefits of Rot6D during training, our results show that the same property is important for real-world transfer, possibly because discontinuous representations induce unstable mappings that overfit to simulation-specific correlations and do not generalize on hardware.

Next, we assess the impact of adding temporal history to the observation. While prior work finds it beneficial~\cite{hwangbo2019learning}, incorporating history degrades performance in our setting. We hypothesize that this results from our minimal domain randomization: additional history may encourage the policy to memorize simulation-specific state–action patterns, leading to a larger distribution shift in real-world dynamics.

We further analyze the sensitivity of sim-to-real performance to the mechanical parameters used in simulation, with armature as an example. While some practitioners modify armature values for numerical stability, we find that using inaccurate parameters noticeably degrades tracking accuracy. In particular, ignoring armature entirely (setting armature to zero) leads to excessive joint accelerations, resulting in overswinging and self-collisions during dynamic motions. More details on armature computation are presented in S1. 

Finally, we investigate the effect of latency in the deployment pipeline. By injecting artificial delays, we find that even small delays can severely impair tracking: a 2 ms delay increases velocity error, a 5 ms delay causes one failure, and a 10 ms delay causes failures in two out of three trials. While randomizing delay during training can potentially mitigate this issue, it also makes learning more difficult. These results highlight the importance of a carefully engineered, real-time deployment framework for robust sim-to-real transfer.

\subsubsection*{Adpative Sampling}
We ablate adaptive sampling during training to evaluate its effectiveness, using convergence speed as the primary metric. As shown in Fig.~\ref{fig:ablation}B, adaptive sampling proves critical for learning difficult segments within long motion sequences. Without it, three out of four motions fail in certain challenging segments even after 30k training iterations---for example, the cartwheels in Motion~1 (Fig.~\ref{fig:ablation}C). For easier motions, such as Motion~4, adaptive sampling also halves the required iterations (2k vs.\ 4k). These results demonstrate that adaptive sampling substantially accelerates training and improves robustness.

To further illustrate its mechanism, we visualize the sampling process in Fig.~\ref{fig:ablation}D. Training begins with a uniform sampling distribution, but as difficult segments cause higher failure rates---such as the two cartwheels in Motion~1---the distribution gradually shifts to favor those segments. This reweighting reduces the variance of the policy gradient and promotes faster convergence. As the policy improves and failure rates decrease in these segments, the distribution becomes smoother and more balanced across all segments. 

\subsubsection*{Latent Diffusion}
To evaluate the contribution of our latent space formulation, we conducted an ablation study using cartwheels, a challenging athletic motion requiring precise spatiotemporal coordination. We compared success rates in sim-to-sim transfer using MuJoCo~\cite{mujoco}, where success was defined as completing the full cartwheel motion without falling.

\begin{figure}
    \centering
    \includegraphics[width=1.0\textwidth]{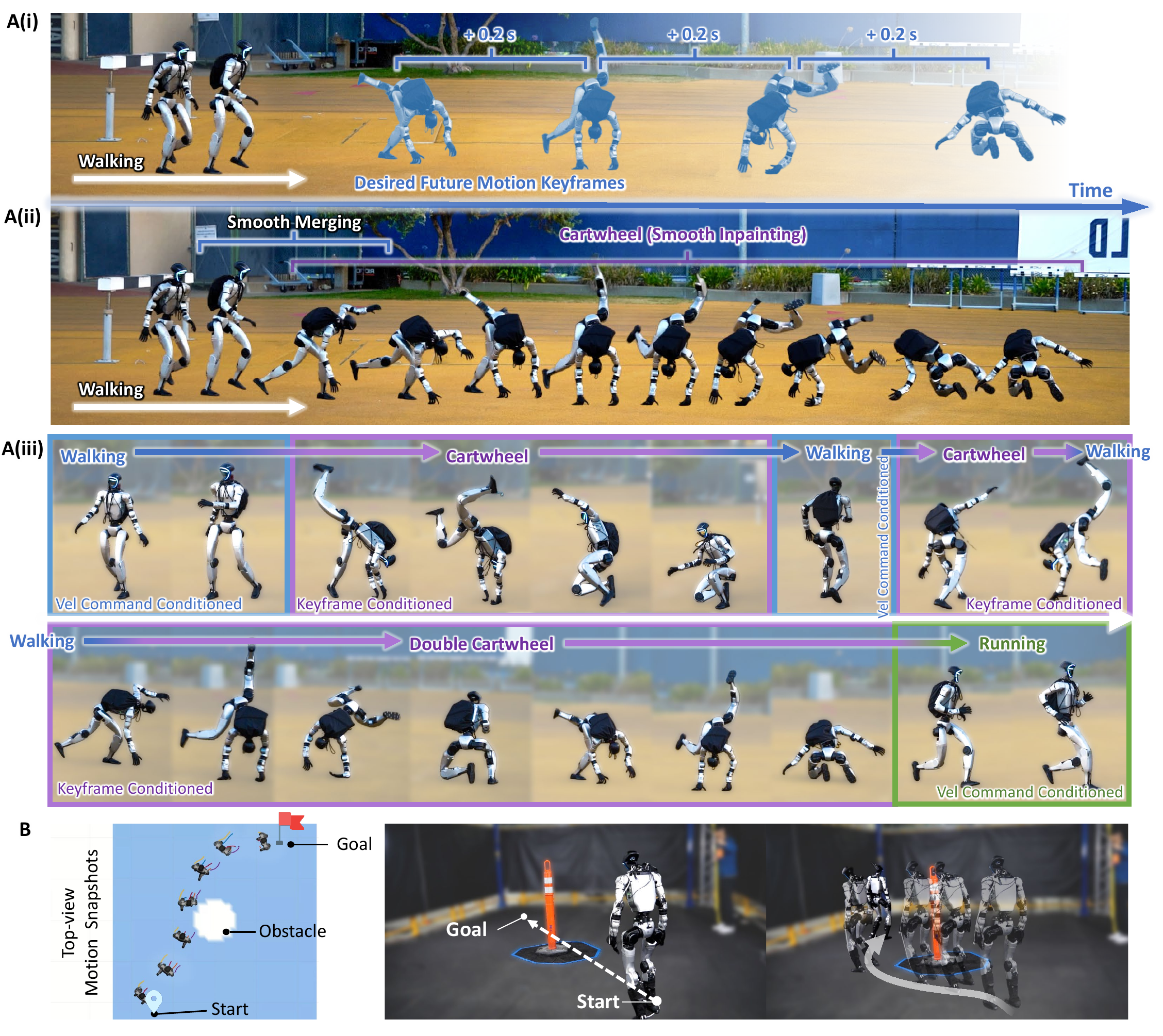}
    \caption{\textbf{Task transition and composition.}
    (\textbf{A}) Motion inpainting with future keyframes. (i) Desired future keyframes with $0.2$~s intervals. (ii) Starting from walking, the robot smoothly completed the cartwheel by inpainting both the transition and the intermediate motion between the keyframes. (iii) Long-horizon execution with four keyframed-conditioned cartwheels interwoven with velocity-conditioned walking and running, achieving smooth multi-round transitions between different tasks and motions, demonstrating its versatility in task specifications.
    (\textbf{B}) Real-world obstacle avoidance. We demonstrated scene-aware navigation by composing waypoint and obstacle-avoidance costs. When given a goal directly ahead, the robot successfully detoured around the obstacle and reached the target.}
    \label{fig:versatile_inpaint_sdf}
\end{figure}

\begin{figure}
    \centering
    \vspace{-0.5cm}
    \includegraphics[width=1.0\textwidth]{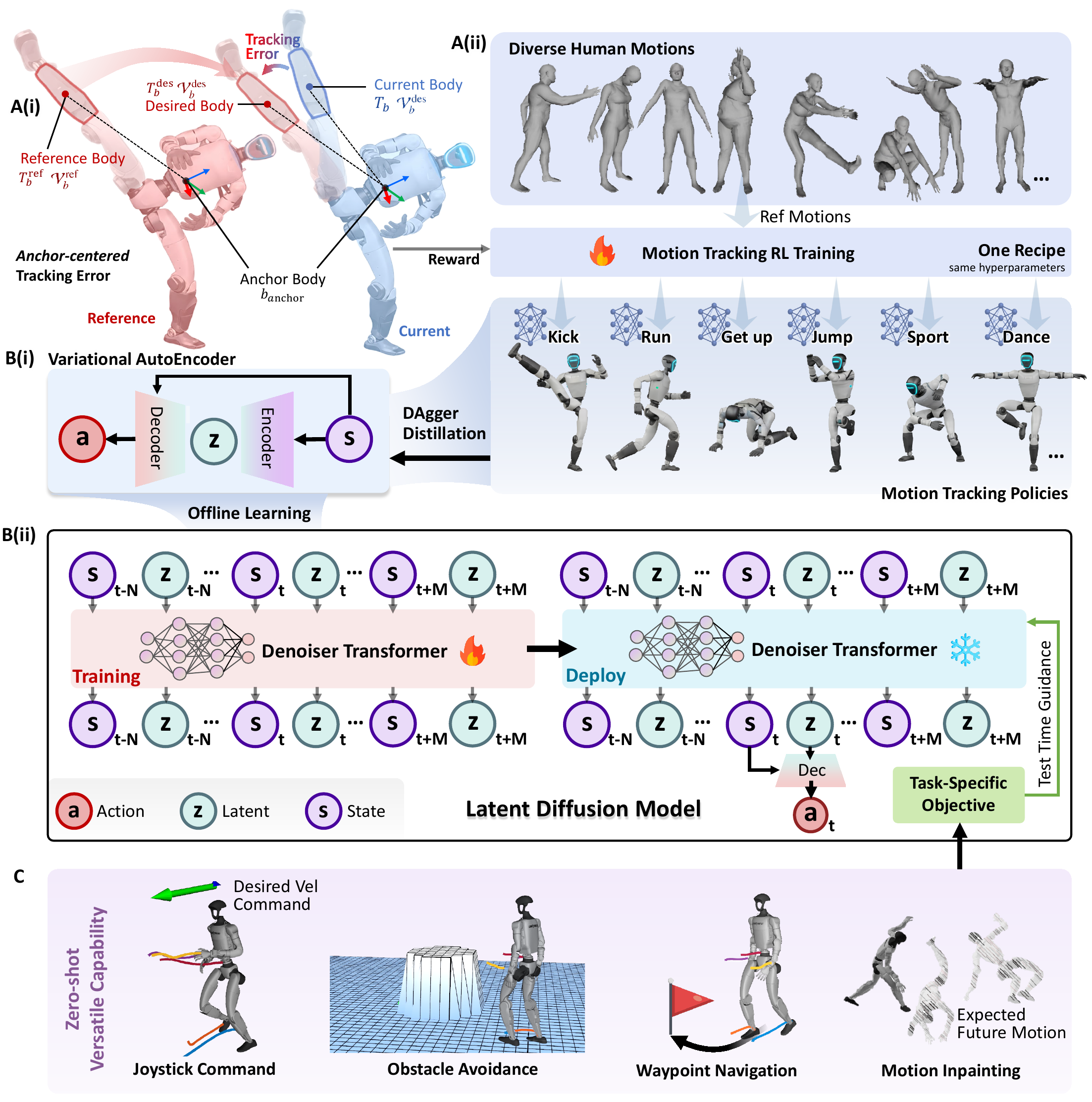}
    \caption{\textbf{Overview of the framework.} (\textbf{A}) Scalable learning of diverse human motions via motion tracking. 
    (i) The target motion is re-anchored to the current pose to allow drift and recovery, which is critical for successful sim-to-real transfer. 
    (ii) Diverse motions are tracked with RL using a single recipe and shared hyperparameters, showing scalability across a broad spectrum of skills. 
    (\textbf{B}) Versatile control via latent state-action diffusion model. (i) Stage 1: A VAE is trained via DAgger to compress the diverse motion tracking policies into a smooth, structured latent space. (ii) Stage 2: A state-latent diffusion model is trained on VAE trajectory rollouts. During inference, the current action is decoded from the denoised latent. (\textbf{C}) Versatile capabilities demonstrated zero-shot on unseen downstream tasks through test-time guidance. 
     }
    \label{fig:pipline}
\end{figure}

\begin{figure}
    \centering
    \includegraphics[width=1.0\textwidth]{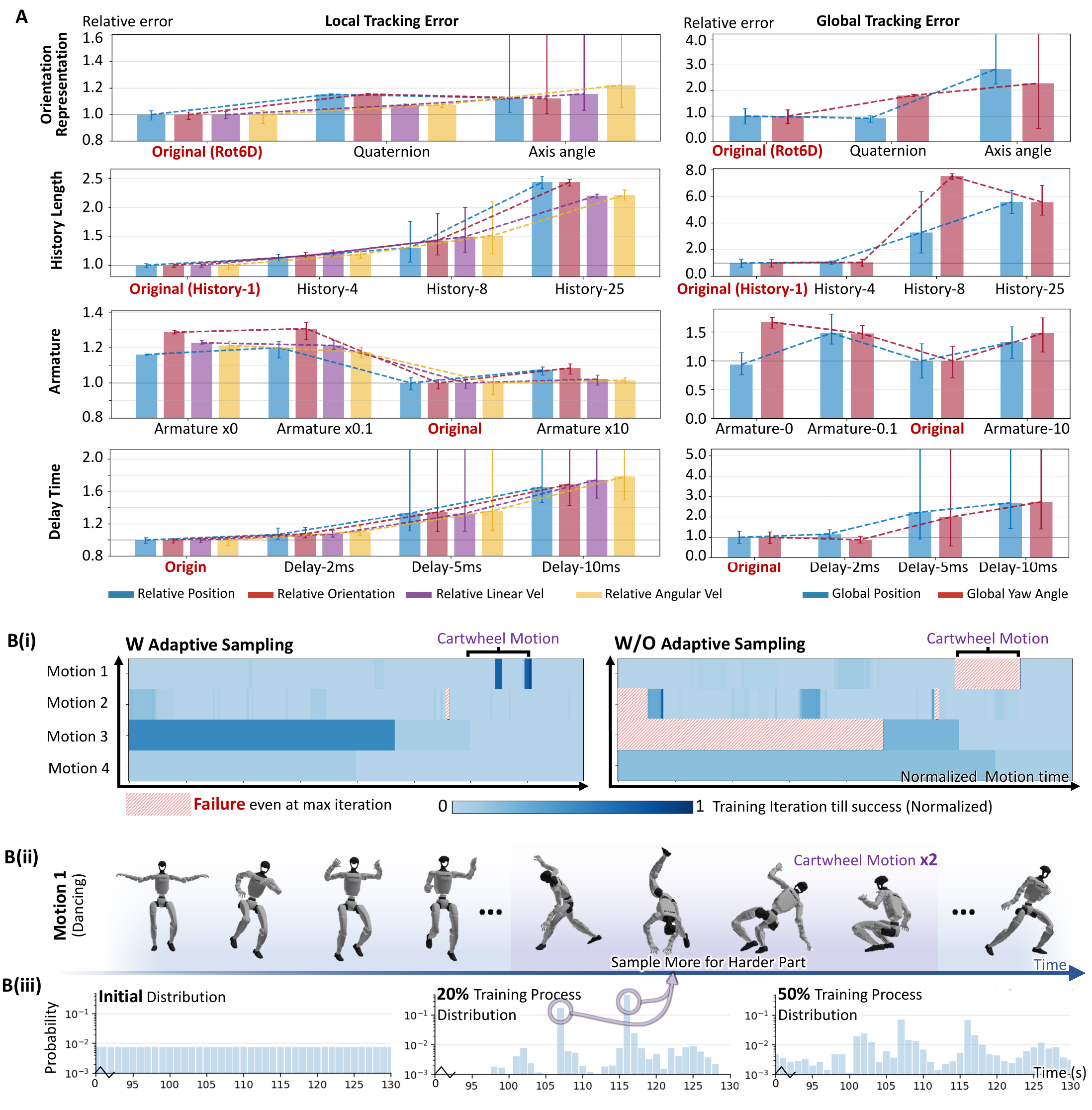}
    \caption{\textbf{Ablation studies.}
    (\textbf{A}) Motion tracking error comparison ablating orientation representation, history length, armature, and delay, showing that continuous orientations, no observation history, correct armature settings, and minimal deployment delay are critical for sim-to-real transfer in our setup. 
    (\textbf{B}) Training performance ablating adaptive sampling (AS). 
    (i) Iterations required to solve the motion; without AS, difficult segments remain unsolved even after 30k iterations.
    (ii) Visualization for \textit{Motion~1}, highlighting two challenging cartwheel segments that cause the baseline to fail.
    (iii) Evolution of the sampling distribution, showing that AS concentrates sampling on difficult segments early on and spreads out as learning progresses.}
    \label{fig:ablation}
\end{figure}

The baseline method without latent encoding achieved only 5\% success rate, whereas our latent diffusion model achieved 95\% success in simulation. Critically, this high performance transferred to the physical robot, as shown in (Fig.~\ref{fig:versatile_inpaint_sdf}). This improvement likely stems from enhanced robustness to variations and noise, consistent with prior work on using latent representations~\cite{agon_vmp}.

\newpage


\clearpage
\bibliography{bibliography}
\bibliographystyle{sciencemag}


\section*{Acknowledgments}
We would like to thank Amazon for letting us use their facility. We would also like to thank Pieter Abbeel, Kevin Zakka, Bike Zhang, Junfeng Long, Zhongyu Li for the thoughtful discussion, Sangli Teng, Zhi Su, and Yiyang Shao for their help in the experiments. 

\paragraph*{Funding:}
This work is supported in part by NSF CMMI-1944722, The Robotics and AI Institute, BAIR Humanoid Intelligence Center, the program ``Design of Robustly Implementable Autonomous and Intelligent Machines (TIAMAT)", Defense Advanced Research Projects Agency award number HR00112490425, and in part by Stanford Institute for Human-Centered AI, Wu-Tsai Human Performance Alliance, NSF FRR-2153854.

\paragraph*{Author contributions:}
Q.L., T.E.T., and X.H. led the project, contributed to the algorithm, software design, experiments, data analysis, and manuscript writing. Y.G. contributed to the experiments, data analysis, and manuscript writing. G.T. contributed to the manuscript writing, guided the project, and discussion. 
K.S. and C.K.L. advised the project and directed the research.
\paragraph*{Competing interests:}
There are no competing interests to declare.

\paragraph*{Data and materials availability:}
All data needed to evaluate the conclusions in the paper are present in the paper of Supplementary Materials and in an online dataset \cite{liao_2025_17529720}.


\subsection*{Supplementary materials}
Supplementary Sections S1 to S4\\
Figs. S1 to S2\\
Tables S1 to S6\\
References \textit{(33, 34, 45, 83-\arabic{enumiv})}\\ 
Movie S1 to S6


\newpage


\renewcommand{\thefigure}{S\arabic{figure}}
\renewcommand{\thetable}{S\arabic{table}}
\renewcommand{\theequation}{S\arabic{equation}}
\renewcommand{\thepage}{S\arabic{page}}
\setcounter{figure}{0}
\setcounter{table}{0}
\setcounter{equation}{0}
\setcounter{page}{1} 


\begin{center}
\section*{Supplementary Materials for\\ \scititle}

\author{
	Qiayuan~Liao$^{1\ast\dagger}$,
	Takara~E.~Truong$^{2\dagger}$,
	Xiaoyu~Huang$^{1\dagger}$, \\
    Yuman Gao$^{1}$,
	Guy~Tevet$^{2}$,
	Koushil~Sreenath$^{1\ddagger}$,
	C.~Karen~Liu$^{2\ddagger}$\and \\
	\small$^{1}$University of California, Berkeley, CA 94720, USA. \\
	\small$^{2}$Stanford University, Stanford, CA 94305, USA.\and \\
	\small$^\ast$Corresponding author. Email: qiayuanl@berkeley.edu\and \\
	\small$^\dagger$These authors contributed equally to this work; order decided by coin toss.\and \\
	\small$^\ddagger$Equal advising; order mirrors the coin toss, with the other lab listed last.
}

\end{center}

\subsubsection*{This PDF file includes:}
Supplementary Sections S1 to S4\\
Figures S1 to S2\\
Tables S1 to S6\\
References \textit{(33, 34, 45, 83-\arabic{enumiv})}\\ 

\subsubsection*{Other Supplementary Materials for this manuscript:}
Movies S1 to S6\\
Dataset \cite{liao_2025_17529720}.

\newpage


\subsection*{S1 Motion Tracking Formulation}

\subsubsection*{Armature}
In a geared actuator, part of the motor’s torque accelerates the motor and transmission themselves, and the remainder is transmitted through the gear train to the joint. Viewed from the joint side, the motor’s rotor inertia appears as an additional inertia at the joint equal to the rotor inertia multiplied by the square of the gear ratio. Here, the gear ratio is defined as motor revolutions per joint revolution. Many simulators refer to this reflected quantity as the \textit{armature}. Crucially, it is a real physical contribution to the joint’s inertia, not a numerical stability knob; setting it incorrectly alters the robot’s apparent inertia and dynamic response rather than merely the simulator’s integrator behavior. Below, we detail how we compute the armature values for the four Unitree G1 robots.

As shown in Table~\ref{tab:actuator_inertia}, the Unitree G1 uses four two-stage planetary actuators. For completeness, we include the stage inertias in the calculation, but they contribute only a small residual--in all cases, the rotor-reflected inertia dominates the actuator armature. Building on Table~\ref{tab:actuator_inertia}, we map each actuator’s armature to the joints summarized in Table~\ref{tab:joint_armature}. Most joints are single-actuated, so their joint armature equals the actuator armature. The exceptions are the ankle roll/pitch and waist roll/pitch joints, which are dual-actuated via a spatial linkage; in the nominal pose, this linkage is effectively 1:1, so we model their joint armature as twice the actuator value. Table~\ref{tab:joint_armature} shows that the reflected inertia contributes a large—often dominant—fraction of the effective axis inertia in the nominal pose, especially toward the distal joints. Practically, this means controller bandwidth and gain selection at these joints are constrained primarily by the actuator’s reflected inertia rather than by the link-subtree inertia.

\subsubsection*{Anchor}
For non-anchor bodies $b\in \mathcal{B}\setminus \{b_\text{anchor}\}$, the desired pose $T_b^{\text{des}} = (\mathbf{p}_b^\text{des}, R_b^\text{des}) = \mathcal A\!\left(T_b^{\text{ref}},\, T_{\text{anchor}}\right)$ is computed as $ R_b^\text{des} = R_{\Delta}R_{b}^\text{ref}$, $ \mathbf{p}_b^\text{des} =\mathbf{p}_\Delta + R_\Delta(\mathbf{p}_{b}^\mathrm{ref} - \mathbf{p}_{\text{anchor}}^\mathrm{ref})$, where $\mathbf{p}_\Delta = [\mathbf{p}_{\text{anchor}.x},\ \mathbf{p}_{\text{anchor}.y},\ \mathbf{p}_{\text{anchor}.z}^{\mathrm{ref}}]$ and $R_\Delta = R_z(\operatorname{yaw}(R_{\text{anchor}} {R^\mathrm{ref}_{\text{anchor}}}^\top))$. It is a transform shifts the motion into the robot's local frame by preserving height, aligning yaw, and translating the $xy$ origin under the robot.

\subsubsection*{Reward}
To calculate the average tracking error, for each body $b \in \mathcal{B}_\text{target}$, we compute the position, orientation, linear and velocity tracking errors between the desired and actual poses and twists as  
$\mathbf{e}_{p,b} = \mathbf{p}_b^\text{des} - \mathbf{p}_b$, 
$\mathbf{e}_{R,b} = \log(R_b^\text{des} R_b^\top)$, 
$\mathbf{e}_{v,b} = \mathbf{v}_b^\text{des} - \mathbf{v}_b$, and 
$\mathbf{e}_{\omega,b} \approx \boldsymbol{\omega}_b^\text{des} - \boldsymbol{\omega}_b$, 
assuming the orientation error is small. 
The mean squared errors are then averaged over all target bodies as 
$\bar{e}_{s} = \tfrac{1}{|\mathcal{B}_\text{target}|} \sum_{b \in \mathcal{B}_\text{target}} \|\mathbf{e}_{s,b}\|^2$, 
where $s \in \{\mathbf{p}, R, \mathbf{v}, \boldsymbol{\omega}\}$. These errors are then normalized by a Gaussian-shaped exponential function: 
$r(\bar{e}_{s},\sigma_s) = \exp\left(-{\bar{e}_{s}}/{\sigma^2_s}\right)$, where $\sigma$ for each term can be seen as its tolerance scale, and we choose values that fit most motions. The nominal error for each tracking term is as follows: $\sigma_p = 0.3$ for position, $\sigma_R = 0.4$ for orientation, $\sigma_v = 1.0$ for linear velocity, and $\sigma_\omega = 3.14$ for angular velocity.

The formulations for the regularization terms are as follows: 
$r_{\text{smooth}} = \|\mathbf{a}_t - \mathbf{a}_{t-1}\|_2$ penalizes abrupt changes in consecutive actions, 
$r_{\text{limit}} = \sum_{j=1}^{N_j} [\max(l_j - \theta_j, 0) + \max(\theta_j - u_j, 0)]$ penalizes joint positions $\theta_j$ exceeding the soft limits $[l_j, u_j]$, set to $0.9$ times the mechanical joint limits, 
and $r_{\text{contact}} = \sum_{b \notin \mathcal{B}_{\text{ee}}} \mathbf{1}\!\left[\|\mathbf{f}_b^{\text{self}}\| > f_{\text{th}}\right]$ penalizes excessive self-contact forces, where $f_{\text{th}} = 1\text{N}$.

The weights for each term are set as follows: 
$\lambda_{l} = -10.0$, and $\lambda_{s} = \lambda_{c} = -0.1$. 
If the global tracking term $r_{g}$ is included, its weight is $\lambda_{g} = 0.5$. A summary of the reward terms is provided in Table~\ref{tab:rewardterms}. 

\subsubsection*{Observation}
When position drift compensation is unnecessary or reliable state estimation is unavailable, the \emph{linear} components may be omitted (i.e., the translational part of $\mathbf{e}_{\text{anchor}}$ and the linear component of $\mathcal{V}_{\text{root}}$).
We use asymmetric actor–critics to boost training efficiency. In addition to the policy observation, the critic also receives per-body relative poses w.r.t.\ the anchor, $T^{-1}_{\text{anchor}}\,T_{b},\ \forall b \in \mathcal{B}$, enabling it to estimate body-wise tracking errors directly in Cartesian space.

\subsubsection*{Heuristically Designed Paramters}
We heuristically set stiffness and damping for each joint $j$ following: $k_{\mathrm{p},j} = I_j \omega^2, \quad k_{\mathrm{d},j} = 2 I_j \zeta \omega$, where $\omega$ is the natural frequency, $\zeta$ is the damping ratio, and $I_j = k_{\mathrm{g},j}^2 I_{\text{rotor},j}$ is the reflected inertia (aramature) of the joint. Here, $I_{\text{rotor},j}$ is the rotor inertia and $k_\mathrm{g}$ is the gear-ratio. The reason we consider only the reflected inertia is that the subtree inertia varies in different configurations, and we do not rely on a heuristic to obtain a reasonable value; exact precision is not required. Note that these heuristically are built in function simulation like MuJoCo and IsaacSim for solver stability, and are used~\cite{zordan2002motion, zordan2005dynamic, farbod2025rss}.
We choose a damping ratio $\zeta = 2$ (overdamped) rather, since the inertia is typically underestimated by considering only the motor armature and ignoring the apparent link inertia. The natural frequency is set to a relatively low value of $10\,\text{Hz}$, which promotes compliance with moderate gains.

We use $\boldsymbol{\alpha} = 0.25\frac{\boldsymbol{\tau}_{\text{max}}}{k_{\mathrm{p},j}}$, with $\boldsymbol{\tau}_{j,\text{max}}$ representing the maximum allowable joint torque for joint $j$. This heuristic assumes that contacts generally occur around $\boldsymbol{\theta}^{0}$ and that the robot hardware design ensures the maximum joint torque is proportional to the expected load.

\subsubsection*{Domain Randomization}
Here we provide the parameters for domain randomization. All ranges below denote independent uniform draws at the start of each training episode unless noted otherwise.

We randomize the robot’s contact friction and restitution across all bodies, sampling static friction 
$\mu_\mathrm{static} \sim \mathcal{U}(0.3,\,1.6)$, 
dynamic friction $\mu_\mathrm{dynamic} \sim \mathcal{U}(0.3,\,1.2)$, 
and restitution $e_\mathrm{rest} \sim \mathcal{U}(0.0,\,0.5)$. 
To simulate joint calibration offsets that affect both actions and observations, the default joint positions are perturbed additively by 
$\Delta\theta^0_j \sim \mathcal{U}(-0.01,\,0.01)\,\mathrm{rad}$ 
for most joints and by a larger range 
$\Delta\theta^0_j \sim \mathcal{U}(-0.1,\,0.1)\,\mathrm{rad}$ 
for ankle joints to account for greater calibration errors. 
The torso’s center of mass is randomized along all three axes, with 
$\Delta x \sim \mathcal{U}(-0.025,\,0.025)\,\mathrm{ref}$, 
$\Delta y \sim \mathcal{U}(-0.05,\,0.05)\,\mathrm{ref}$, 
and $\Delta z \sim \mathcal{U}(-0.05,\,0.05)\,\mathrm{ref}$. 
To further enhance robustness to external disturbances, random velocity perturbations are applied periodically during training, 
with intervals $\Delta t \sim \mathcal{U}(1.0,\,3.0)\,\mathrm{s}$. 
The perturbation magnitudes are uniformly sampled from translational velocity ranges 
$(x, y, z) \in \{[-0.5, 0.5], [-0.5, 0.5], [-0.2, 0.2]\}\,\mathrm{m/s}$ 
and rotational velocity ranges (roll, pitch, yaw) 
$\in \{[-0.52, 0.52], [-0.52, 0.52], [-0.78, 0.78]\}\,\mathrm{rad/s}$. A summary of the domain randomization terms is provided in Table~\ref{tab:domain_rand}.

\subsubsection*{Termination and Adaptive Sampling}
An episode terminates when the tracking error becomes unrecoverably large. 
The position and orientation errors are defined as 
$\mathbf{e}_{p,b} = \mathbf{p}_{b}^\mathrm{d} - \mathbf{p}_b$ 
and 
$\mathbf{e}_{R,b} = \log(R_{b}^\mathrm{d} R_b^\top)$. 
Termination occurs when either the anchor body $b_\text{anchor}$ or any end-effector body 
$b \in \mathcal{B}_\text{ee} = \{\text{left/right ankles, left/right hands}\}$ 
deviates excessively from the reference, that is, when 
$|\mathbf{e}_{p,z,b}| > 0.25~\text{m}$ or $\|\mathbf{e}_{R,\text{anchor}}\| > 0.8~\text{rad}$. 

At each episode reset, the motion phase is adaptively sampled from the reference trajectory 
to balance learning across segments of varying difficulty. 
The reference motion is divided into $S$ one-second bins, and each bin’s failure rate is updated using an exponential moving average, $\bar{f}_s \leftarrow 0.999\,\bar{f}_s + 0.001 f_s$, where $\bar{f}_\text{s}$ is the smoothed failure rate, to mitigate short-term fluctuations. 
The resulting failure rates are offset by a small uniform floor, $\bar{f}_s = \bar{f}_s + 0.1/S$, so that when all bins have uniformly low failure rates, the sampler naturally reverts to uniform sampling. 
Since failures are more likely caused by suboptimal actions taken shortly before termination, we apply a non-causal convolution with an exponentially decaying kernel 
$k(u)=\rho^{u}$ with $\rho=0.8$ and a look-back window of $u \in \{0,1,2\}$. 
The sampling probability of each bin $\text{s}$ with the convolution kernel is then computed as
\[
p_\text{s} = 
\frac{\sum_{u=0}^{K-1} \rho^u\, \bar{f}_{\text{s}+u}}
{\sum_{j=1}^\text{S} \sum_{u=0}^{K-1} \rho^u\, \bar{f}_{j+u}}.
\]
The probabilities are then normalized as $\hat{p}_s = p_s / \sum_{j=1}^{S} p_j$. 
Finally, the bin of the starting phase is drawn from $\mathrm{Multinomial}(\hat{p}_1,\dots,\hat{p}_S)$.

The robot is then initialized at the corresponding reference configuration and velocity for the sampled phase, 
with small random perturbations applied to counteract cumulative errors during long-horizon rollouts. Specifically, pose perturbations are sampled within 
$x,y \in [-0.05, 0.05]~\text{m}$, 
$z \in [-0.01, 0.01]~\text{m}$, 
$\text{roll}, \text{pitch} \in [-0.1, 0.1]~\text{rad}$, 
and $\text{yaw} \in [-0.2, 0.2]~\text{rad}$, 
while velocity perturbations follow the same range used in the domain randomization settings.

\subsection*{S2 Motion Tracking Results}
\subsubsection*{Complete List of Validated Motions}
We used various datasets to showcase the wide applicability of our framework. These include datasets from prior works~\cite{he2025asap, zhang2025hub}, Unitree-retargeted LAFAN1~\cite{lafan1}, and online animation data\footnote{Motion sources: Reallusion “MD Panther Lady” pack (\url{https://www.reallusion.com/ContentStore/iClone/3d-animation/panther-lady/default.html}) and “Martial Arts – Taekwondo” motion pack (\url{https://actorcore.reallusion.com/3d-motion/pack/martial-arts-taekwondo}).}. 
The complete list of motions validated in simulation and on hardware is provided in Table~\ref{tab:skill_success}. 

\subsubsection*{Extra Ablation on PD Gains}
We ablate the PD gains of the impedance controller. As described earlier, the gains are parameterized by a natural frequency $\omega$, where a larger $\omega$ results in a stiffer response. We evaluate several values of $w$ and also compare against the gains used in ASAP~\cite{he2025asap}. As shown in Fig.~\ref{fig:extra-ablation}, $\omega = 10 \text{Hz}$ achieves the best global tracking performance and stronger local tracking than both $\omega = 5~\text{Hz}$ and ASAP. Although $\omega = 25~\text{Hz}$ yields slightly lower local pose error, the higher stiffness produces noticeable high-frequency oscillation and loud sound in the actuator gearboxes, which likely results from the large torque overshoots under high gains. Therefore, to balance performance and hardware lifespan, we adopt $\omega = 10~\text{Hz}$ in all experiments.

\subsection*{S3 Diffusion Formulation}

\subsubsection*{Diffusion State}

We use character frames to normalize the states for consistency across trajectories. A \textit{character frame} \( \mathcal{C}_{t'} \) is defined for each timestep \( t' \) 
by the root position \( \mathbf{p}_{\text{root}}^{\,t'} \) and yaw rotation 
\( \mathbf{R}_{\text{yaw}}^{\,t'} \) (with pitch and roll removed). 
All quantities expressed in \( \mathcal{C}_{t'} \) are thus invariant to global translation and yaw rotation. 
At the \textit{current timestep} \( t \), we define the frame \( \mathcal{C}_t \) 
as the current character frame for trajectory normalization. For each timestep \( t + n\)  within the history and horizon 
(\( n \in [-N, H] \)), the state \(\mathbf{s}_{t+n}\) is consisted of root features and body features. 

The root features of \(\mathbf{s}_{t+n}\) are expressed relative to the current frame \( \mathcal{C}_t \), i.e., the root pose and twist are computed as
\begin{equation}
\mathbf{p}_{\text{root}}^{\,\text{rel}}(t{+}n)
= (\mathbf{R}_{\text{yaw}}^{\,t})^\top
\!\big(\mathbf{p}_{\text{root}}^{\,t+n}-\mathbf{p}_{\text{root}}^{\,t}\big),
\qquad
\mathbf{R}_{\text{root}}^{\,\text{rel}}(t{+}n)
= (\mathbf{R}_{\text{yaw}}^{\,t})^\top \mathbf{R}_{\text{root}}^{\,t+n},
\end{equation}
\begin{equation}
\mathbf{v}_{\text{root}}^{\,\text{rel}}(t{+}n)
= (\mathbf{R}_{\text{yaw}}^{\,t})^\top
\!\big(\mathbf{v}_{\text{root}}^{\,t+n}-\mathbf{v}_{\text{root}}^{\,t}\big),
\qquad
\boldsymbol{\omega}_{\text{root}}^{\,\text{rel}}(t{+}n)
= (\mathbf{R}_{\text{yaw}}^{\,t})^\top \boldsymbol{\omega}_{\text{root}}^{\,t+n}.
\end{equation}

The body features of state \(\mathbf{s}_{t+n}\) are instead expressed in their 
\textit{local root frames} \( \mathcal{C}_{t+n} \). 
For each body \( b \in \mathcal{B}_\text{target} \), the local positions and velocities are computed as
\begin{equation}
\mathbf{p}_{b}^{\,\text{local}}(t{+}n)
= (\mathbf{R}_{\text{yaw}}^{\,t+n})^\top
\!\big(\mathbf{p}_{b}^{\,t+n}-\mathbf{p}_{\text{root}}^{\,t+n}\big),
\qquad
\mathbf{v}_{b}^{\,\text{local}}(t{+}n)
= (\mathbf{R}_{\text{yaw}}^{\,t+n})^\top
\!\big(\mathbf{v}_{b}^{\,t+n}-\mathbf{v}_{\text{root}}^{\,t+n}\big).
\end{equation}

This hybrid trajectory representation expresses root motion in the current 
character–yaw frame and body motion in their instantaneous local frames, encoding temporal information invariant to global translation and yaw while retaining detailed local structure that eases model learning. 

To enhance the temporal information in the state representation, we integrate an \textit{emphasis projection}~\cite{Guided-MDM,Diffuse-CLoC}. 
The projection matrix is defined as \( \mathbf{P} = [\mathbf{A}\mathbf{B} \ \ \mathbf{I}]^\top \), 
where \( \mathbf{A}_{ij} \sim \mathcal{N}(0, 1) \) is a random Gaussian matrix 
and \( \mathbf{B} \) is a diagonal matrix with entries set to \( c = 6 \) 
for the dimensions corresponding to root poses and twists. 
This formulation again jointly encodes root features with temporal information and local body features, with the former being emphasized by the projection.
The transformed input to the latent diffusion model is then \( \mathbf{s}' = \mathbf{P}\mathbf{s} \), 
and after the LDM predicts the next state \( \hat{\mathbf{s}}' \) in the projected space, the corresponding state in the original space is recovered via the inverse transformation 
\( \hat{\mathbf{s}} = \mathbf{P}^{-1}\hat{\mathbf{s}}' \), where \( \mathbf{P}^{-1} \) is computed using the pseudoinverse. 

\subsubsection*{Diffusion Dataset Collection}

Because the diffusion model is trained entirely on offline data, using only clean VAE rollouts often causes out-of-distribution behavior during execution. To enhance robustness, we follow PDP~\cite{takara_PDP} to augment trajectories with policy perturbations, forming an error band that enables the model to recover from deviations during deployment. Specifically, we roll out the VAE policies with added action noise while recording the original states and latents. 

Prior works~\cite{takara_PDP, Diffuse-CLoC} add i.i.d. Gaussian action noise at each step, but the overdamped PD gains suppress such high-frequency perturbations, limiting state diversity. 
We instead use Ornstein-Uhlenbeck (OU) noise~\cite{Uhlenbeck}, which produces temporally correlated action perturbations:
\begin{equation}
\eta_{t+1} = \eta_t + \theta (\mu - \eta_t)\, \Delta t + \sigma \sqrt{\Delta t}\, \varepsilon_t,
\quad \varepsilon_t \sim \mathcal{N}(0, I),
\end{equation}
where $\theta = 0.8$ controls the mean reversion rate, $\mu = 0$ is the long-term mean, and $\Delta t = 1.0$. 
We set a joint-wise noise scale of $\sigma = 0.1$. 
Actions are then perturbed as $a_t \leftarrow a_t + \eta_t$, creating a persistent \emph{error band} that the policy learns to recover from during deployment. Episodes are collected such that each sample in the trajectory appears approximately 100 times across the dataset. Each rollout executes the policy for 2.5 seconds but allows the motion to continue for 5 seconds to verify stability. If the robot fails before 5 seconds, the episode is rejected and not included in the dataset.

\subsubsection*{Task Costs}
We show joystick steering, waypoint navigation, and obstacle avoidance as examples of task-specific cost functions. For joystick steering, we define the cost function as the squared difference between the state prediction and the joystick input:

\begin{equation}
    G_{\mathrm{js}}(\hat{\mathbf{\tau}}_t) = \frac{1}{2} \sum_{i=0}^{H} \|V_{xy,i}(\hat{\mathbf{\tau}}_t) - \mathbf{g}_{v} \|^2
\end{equation}
where $V_{xy,i}(\hat{\mathbf{\tau}}_t)$ extracts the predicted planar root velocity at horizon step $i$ and $\mathbf{g}_v \in \mathbb{R}^2$ is the commanded velocity.

For waypoint navigation, the cost transitions from position tracking to velocity minimization near the goal:
\begin{equation}
G_{\mathrm{wp}}(\hat{\mathbf{\tau}}_t) = \sum_{i=0}^{H} (1-e^{-2d_i})\|P_{xy,i}(\hat{\mathbf{\tau}}_t) - \mathbf{g}_{p}\|^2 + e^{-2d_i}\|V_{xy,i}(\hat{\mathbf{\tau}}_t)\|^2
\end{equation}
where $P_{xy,i}(\hat{\mathbf{\tau}}_t)$ extracts planar root position at horizon step $i$, 
$d_i = \|P_{xy,i}(\hat{\mathbf{\tau}}_t) - \mathbf{g}_{p}\|$ is the distance to goal 
$\mathbf{g}_{p} \in \mathbb{R}^2$.

For obstacle avoidance, we use a Signed Distance Field (SDF) with a barrier function:
\begin{equation}
    G_{\mathrm{sdf}}(\hat{\mathbf{\tau}}_t) = \sum_{i=0}^{H} \sum_{b\in\mathcal{B}} B(\mathrm{SDF}(P_{b,i}(\hat{\mathbf{\tau}}_t))-r_b, \delta)
\end{equation}
where $P_{b,i}(\hat{\mathbf{\tau}}_t)$ is the position of body $b$ at horizon step $i$, 
$r_b$ is its collision radius, and $B(x, \delta)$ is a relaxed barrier function \cite{grandia2019feedback}:
\begin{equation}
    B(x, \delta) = \begin{cases}
        -\ln(x) & \text{if } x \geq \delta \\
        -\ln(\delta) + \frac{1}{2} \left[ \left( \frac{x-2\delta}{\delta} \right)^2 - 1 \right] & \text{if } x < \delta
    \end{cases}
\end{equation}

\subsection*{S4 Implementation Details}

\subsubsection*{Motion Tracking}
We use MLPs as our policy networks, and PPO~\cite{schulman2017proximal} as the RL algorithm. Policies run at 50Hz. We provide the architecture and training details for motion tracking in Table \ref{tab:ppo_hyperparameters}. 

We deploy our motion tracking policies, trained using the proposed method, on Unitree G1 humanoid robots. All deployment code is written in C++ and optimized for real-time execution. Full-state estimation is provided at $500$~Hz using a low-level generalized momentum observer combined with a Kalman filter~\cite{flayols2017experimental}. Notably, no external motion capture system is used for tracking.

For extreme, contact-rich behaviors (e.g., getting up from the ground), we either incorporate LiDAR-inertial odometry (LIO)~\cite{koide2024glim} for position correction or exclude state-estimation-dependent observations altogether. All policies are executed onboard using ONNX Runtime~\cite{onnxruntime} on the robot’s CPU. Each inference step takes under $1.0$~ms, allowing seamless integration into the real-time estimation and control loop.

\subsubsection*{Diffusion}
We use MLPs for the VAE encoder and decoder networks, and a Transformer~\cite{vaswani2017attention} Encoder for the diffusion backbone. The transformer has a parameter count of \~19.8M. We train and deploy tracking policies at 25 Hz instead of 50 Hz to allow sufficient time for diffusion model inference, following the exact same formulation as introduced. We provide details for the architecture and training of the VAE in Table \ref{tab:vae_hyperparameters} and Diffusion in Table \ref{tab:diffusion_hyperparameters}. 

In addition, we apply morphological symmetry augmentation in the sagittal plane when learning from the motion-tracking policies. Following the formulation in prior work~\cite{su2024leveraging}, we train both the VAE and the diffusion model on two copies of the data: the original state–action pairs and their sagittal-symmetric counterparts. This effectively doubles the skill repertoire with minimal additional effort. 

At test time, the diffusion policy runs onboard using a portable Mini PC with an NVIDIA RTX 4060 Mobile GPU. We use TensorRT to accelerate the inference speed. The diffusion model inference is performed asynchronously in a separate thread due to latency; each step takes approximately $20\,\mathrm{ms}$ using 20 denoising steps. The light-weight VAE decoder is inferenced in synchronously and on CPU instead.  Gradients of the guiding cost are computed automatically using CppAD~\cite{cppad2019} within each denoising iteration. For the joystick commands and motion inpainting tasks, we use proprioceptive state estimation to provide body pose and velocities. For the waypoint navigation and obstacle avoidance tasks, motion capture data is used to provide both environmental context for cost computation and improved localization to accomplish the tasks.


\newpage


\begin{figure}
    \centering
    \includegraphics[width=1.0\textwidth]{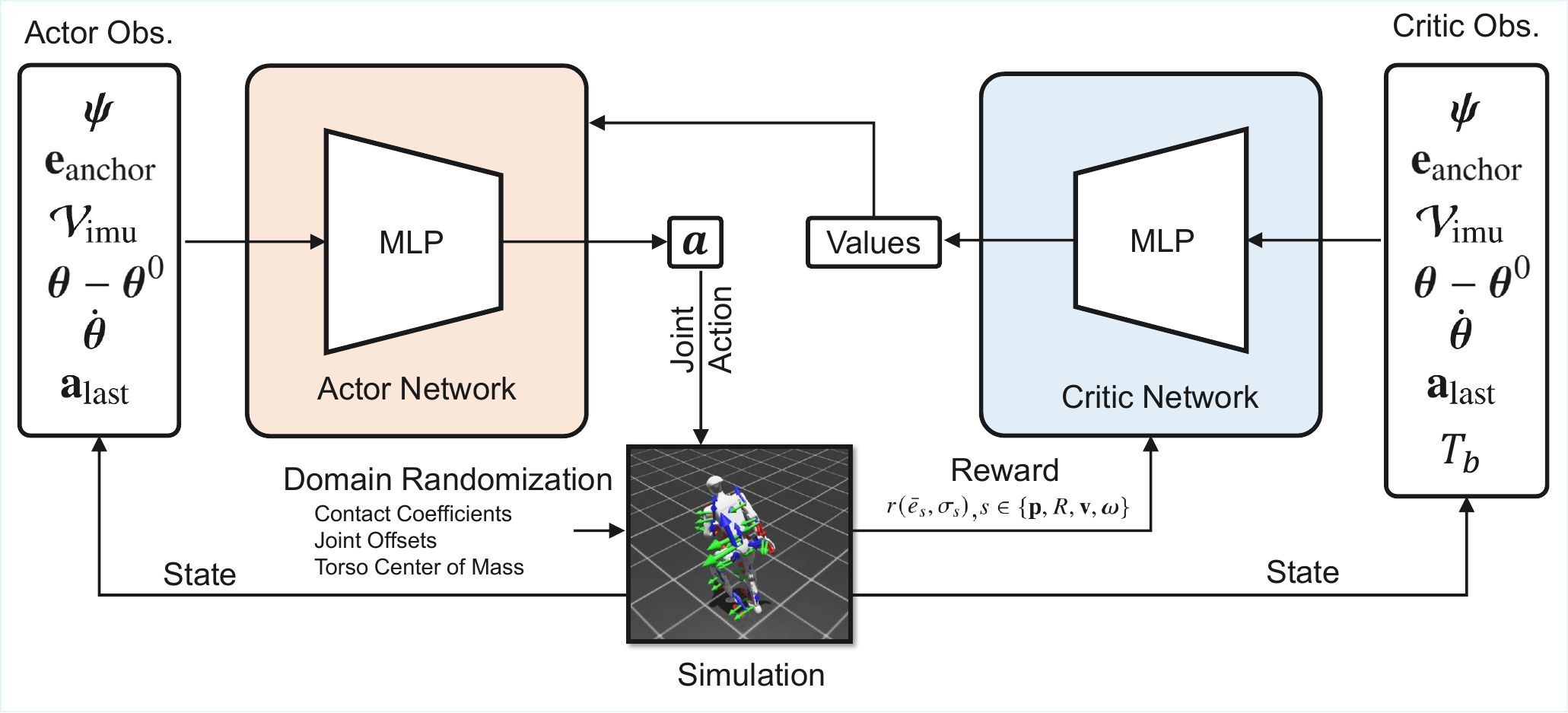}
    \caption{\textbf{Motion Tracking Overview.} Our actor-critic architecture for motion tracking. The actor network takes observations: motion phase $\boldsymbol{\psi}$, anchor pose error $\mathbf{e}_{\text{anchor}}$, IMU twist $\mathcal{V}_{\text{imu}}$, relative joint positions $\boldsymbol{\theta} - \boldsymbol{\theta}^{0}$, joint velocities $\dot{\boldsymbol{\theta}}$, and previous action $\mathbf{a}_{\text{last}}$. The critic receives the same observations in addition to the per-body relative poses w.r.t. the anchor $T_{\text{b}}$ for a desired set of bodies \(b \in \mathcal{B}_\text{target}\). We compute tracking rewards for position $\mathbf{p}$, orientation $R$, linear velocity $\mathbf{v}$, and angular velocity $\boldsymbol{\omega}$ between the desired and actual poses and twists, in addition to three regularization terms. We domain randomize over contact coefficients, joint offsets, and torso center of mass only.}
    \label{fig:tracking_pipeline}
\end{figure}

\begin{figure}
    \centering
    \includegraphics[width=1.0\textwidth]{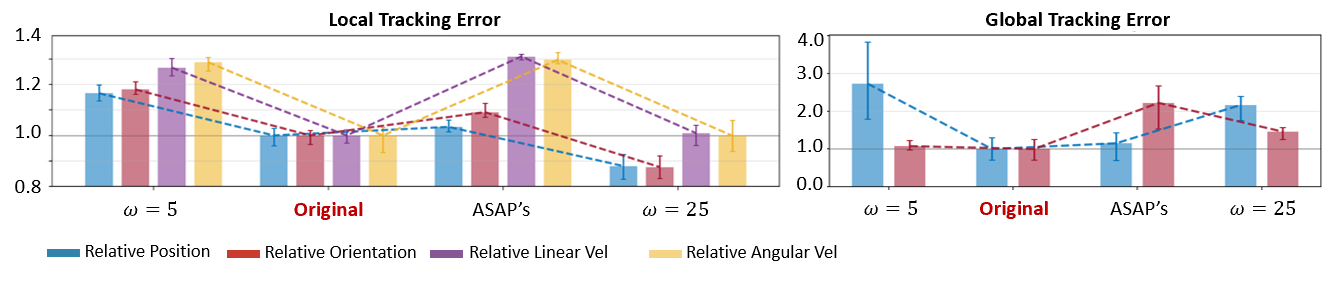}
    \caption{\textbf{Extra Ablation On PD Gains} Ablating on PD Gains, We find that following our heuristics formula with a moderate natural frequency yields the best global tracking performance and surpasses the hand-tuned gains used in prior work~\cite{he2025asap}. Although increasing the natural frequency can further reduce local tracking error, it results in significant torque overshoots that produces audible jitter and can stress the hardware. }
    \label{fig:extra-ablation}
\end{figure}


\begin{table}[t]
\centering
\caption{Unified reward formulation using Gaussian-shaped tracking scores.}
\label{tab:rewardterms}
\begin{tabular}{llc}
\toprule
\textbf{Reward Terms} & \textbf{Equation} & \textbf{Weight} \\
\midrule
\multicolumn{3}{l}{\emph{Task (Tracking)}}\\
Body Position
& $\displaystyle
\exp\!\Big(
-\big( \tfrac{1}{|\mathcal{B}_{\mathrm{target}}|}
\sum_{b\in\mathcal{B}_{\mathrm{target}}}
\|\mathbf{p}^{\mathrm{des}}_{b}-\mathbf{p}_{b}\|^{2} \big) / 0.3^{2}
\Big)$
& $1.0$\\[2mm]
Body Orientation
& $\displaystyle
\exp\!\Big(
-\big( \tfrac{1}{|\mathcal{B}_{\mathrm{target}}|}
\sum_{b\in\mathcal{B}_{\mathrm{target}}}
\|\log(R^{\mathrm{des}}_{b}R_{b}^{\top})\|^{2} \big) / 0.4^{2}
\Big)$
& $1.0$\\[2mm]
Body Linear velocity
& $\displaystyle
\exp\!\Big(
-\big( \tfrac{1}{|\mathcal{B}_{\mathrm{target}}|}
\sum_{b\in\mathcal{B}_{\mathrm{target}}}
\|\mathbf{v}^{\mathrm{des}}_{b}-\mathbf{v}_{b}\|^{2} \big) / 1.0^{2}
\Big)$
& $1.0$\\[2mm]
Body Angular velocity
& $\displaystyle
\exp\!\Big(
-\big( \tfrac{1}{|\mathcal{B}_{\mathrm{target}}|}
\sum_{b\in\mathcal{B}_{\mathrm{target}}}
\|\boldsymbol{\omega}^{\mathrm{des}}_{b}-\boldsymbol{\omega}_{b}\|^{2} \big) / 3.14^{2}
\Big)$
& $1.0$\\[2mm]
Anchor Position (Optional)
& $\displaystyle
\exp\!\Big(
-\|\mathbf{p}^{\mathrm{des}}_{\text{anchor}}-\mathbf{p}_{\text{anchor}}\|^{2} / 0.3^{2}
\Big)$
& $0.5$\\[2mm]
Anchor Orientation (Optional)
& $\displaystyle
\exp\!\Big(
-\|\log(R^{\mathrm{des}}_{\text{anchor}}R_{\text{anchor}}^{\top})\|^{2} / 0.4^{2}
\Big)$
& $0.5$\\[2mm]
\midrule
\multicolumn{3}{l}{\emph{Regularization}}\\
Action smoothness
& $\|\mathbf{a}_{t}-\mathbf{a}_{t-1}\|^{2}$
& $-0.1$\\
Joint position limit
& $\sum_{j=1}^{N}\big[\max(l_j-\theta_j,0)+\max(\theta_j-u_j,0)\big]$
& $-10.0$\\
Undesired self-contacts
& $\sum_{b\notin\mathcal{B}_{\mathrm{ee}}}
\mathbf{1}\!\left[\|f^{\mathrm{self}}_{b}\|>1  \text{N}\right]$
& $-0.1$\\
\bottomrule
\end{tabular}
\end{table}

\begin{table}[t]
\centering
\caption{\textbf{Domain randomization parameters.} ($\mathcal{U}[\cdot]$: uniform distribution)}
\label{tab:domain_rand}
\begin{tabular}{ll}
\toprule
\textbf{Domain Randomization} & \textbf{Sampling Distribution} \\
\midrule
\multicolumn{2}{l}{\emph{Physical parameters}}\\
Static friction coefficients & $\mu_{\text{static}} \sim \mathcal{U}[0.3,\,1.6]$ \\
Dynamic friction coefficients & $\mu_{\text{dynamic}} \sim \mathcal{U}[0.3,\,1.2]$ \\
Restitution coefficient & $e_{\text{rest}} \sim \mathcal{U}[0,\,0.5]$ \\
Default joint positions [rad] & $ \Delta\theta^0_j \!\sim  \mathcal{U}[-0.01,\,0.01] $\\
Torso's COM offset [m]& $\Delta x\!\sim\!\mathcal{U}[-0.025,0.025]$,  $\ \Delta y\!\sim\!\mathcal{U}[-0.05,0.05]$, \\ &$ \Delta z\!\sim\!\mathcal{U}[-0.05,0.05]$ \\
\midrule
\multicolumn{2}{l}{\emph{Root velocity perturbations}}\\
Root linear vel [m/s]& $v_x\!\sim\!\mathcal{U}[-0.5,0.5],\ v_y\!\sim\!\mathcal{U}[-0.5,0.5],\ v_z\!\sim\!\mathcal{U}[-0.2,0.2]$ \\
Push duration [s]& $\Delta t \sim \mathcal{U}[1.0,\,3.0]$ \\
Root angular vel [rad/s]& $\omega_x,\ \omega_y\!\sim\!\mathcal{U}[-0.52,0.52],\ \omega_z\!\sim\!\mathcal{U}[-0.78,0.78]$ \\
\bottomrule
\end{tabular}
\end{table}

\begin{table}[ht]
  \centering
  \caption{\textbf{Actuator reflected inertia.} Total $J_\mathrm{arm} = J_r(g_1g_2)^2 + J_1g_2^2 + J_2$ with unit kg$\cdot$m$^2$. Near-zero entries shown as ``--''.}
  \label{tab:actuator_inertia}
  \begin{tabular}{lcccccccc}
    \\
    \hline
    Actuator & $g_1$ & $g_2$ & $J_r$ & $J_1$ & $J_2$ & $J_r(g_1g_2)^2$ & $J_1g_2^2$ & Total \\
    \hline
    5020-16 & 3.56 & 4.5 & 1.390e-05 & 1.700e-06 & 1.690e-05 & 3.558e-03 & 3.443e-05 & 3.610e-03\\
    7520-14.3 & 4.5 & 3.18 & 4.890e-05 & 9.800e-06 & 5.330e-05 & 1.003e-02 & 9.921e-05 & 1.018e-02\\
    7520-22.5 & 4.5 & 5 & 4.890e-05 & 1.090e-05 & 7.380e-05 & 2.476e-02 & 2.725e-04 & 2.510e-02\\
    4010-25 & 5 & 5 & 6.800e-06 & -- & -- & 4.250e-03 & -- & 4.250e-03\\
    \hline
  \end{tabular}
\end{table}

\begin{table}[ht]
  \centering
  \caption{\textbf{Joint-axis inertia for left arm and left leg.} Columns list $I_\text{axis}^\text{subtree}$ (subtree inertia about the joint axis), $I_\text{armature}$ (reflected motor/gear inertia), $I_\text{axis}^\text{eff} = I_\text{axis}^\text{subtree} + I_\text{armature}$, and the percentage of armature in the effective inertia.}
  \label{tab:joint_armature}
  \begin{tabular}{lcccc}
    \\
    \hline
    Joint & $I_\text{axis}^\text{subtree}$ & $I_\text{armature}$ & $I_\text{axis}^\text{eff}$ & Armature \% \\
     & (kg$\cdot$m$^2$) & (kg$\cdot$m$^2$) & (kg$\cdot$m$^2$) & (\%) \\
    \hline
    \multicolumn{5}{l}{\textit{Left Arm}}\\
    left\_shoulder\_pitch\_joint & 1.748e-01 & 3.610e-03 & 1.785e-01 & 2.0\%\\
    left\_shoulder\_roll\_joint & 1.324e-01 & 3.610e-03 & 1.360e-01 & 2.7\%\\
    left\_shoulder\_yaw\_joint & 2.742e-02 & 3.610e-03 & 3.103e-02 & 11.6\%\\
    left\_elbow\_joint & 3.445e-02 & 3.610e-03 & 3.806e-02 & 9.5\%\\
    left\_wrist\_roll\_joint & 3.664e-04 & 3.610e-03 & 3.976e-03 & 90.8\%\\
    left\_wrist\_pitch\_joint & 4.804e-03 & 4.250e-03 & 9.054e-03 & 46.9\%\\
    left\_wrist\_yaw\_joint & 1.837e-03 & 4.250e-03 & 6.087e-03 & 69.8\%\\
    \hline
    \multicolumn{5}{l}{\textit{Left Leg}}\\
    left\_hip\_pitch\_joint & 8.543e-01 & 1.018e-02 & 8.644e-01 & 1.2\%\\
    left\_hip\_roll\_joint & 6.398e-01 & 2.510e-02 & 6.649e-01 & 3.8\%\\
    left\_hip\_yaw\_joint & 1.296e-01 & 1.018e-02 & 1.398e-01 & 7.3\%\\
    left\_knee\_joint & 1.115e-01 & 2.510e-02 & 1.366e-01 & 18.4\%\\
    left\_ankle\_pitch\_joint & 2.777e-03 & 7.219e-03 & 9.997e-03 & 72.2\%\\
    left\_ankle\_roll\_joint & 3.871e-04 & 7.219e-03 & 7.607e-03 & 94.9\%\\
    \hline
  \end{tabular}
\end{table}

\begin{table}[ht]
  \centering
  \caption{\textbf{Motion Tracking Hyperparameters}}
  \label{tab:ppo_hyperparameters}
  \begin{tabular}{lr}
    \\
    \hline
    Hyperparameter & Value \\
    \hline
    \multicolumn{2}{c}{Architecture} \\
    \hline
    Actor MLP hidden dimensions & [512, 256, 128] \\
    Critic MLP hidden dimensions & [512, 256, 128] \\
    Activation function & ELU \\
    \hline
    \multicolumn{2}{c}{Training} \\
    \hline
    Steps per environment & 24 \\
    Max iterations & 30,000 \\
    Learning rate & $1 \times 10^{-3}$ \\
    Clip parameter & 0.2 \\
    Entropy coefficient & 0.005 \\
    Value loss coefficient & 1.0 \\
    Discount factor ($\gamma$) & 0.99 \\
    GAE $\lambda$ & 0.95 \\
    Desired KL & 0.01 \\
    Learning epochs & 5 \\
    Mini-batches & 4 \\
    \hline
  \end{tabular}
\end{table}

\begin{table}[ht]
  \centering
  \caption{\textbf{VAE Hyperparameters}}
  \label{tab:vae_hyperparameters}
  \begin{tabular}{lr}
    \\
    \hline
    Hyperparameter & Value \\
    \hline
    \multicolumn{2}{c}{{Architecture}} \\
    \hline
    Latent dimension & 32 \\
    Student encoder MLP hidden dimensions & [2048, 1024, 512] \\
    Student decoder MLP hidden dimensions & [2048, 1024, 512] \\
    Teacher hidden MLP hidden dimensions & [512, 256, 128] \\
    Activation function & ELU \\
    \hline
    \multicolumn{2}{c}{{Training}} \\
    \hline
    Learning rate & $5 \times 10^{-4}$ \\
    Accumulated Gradient steps & 15 \\
    KL loss coefficient & 0.01 \\
    \hline
  \end{tabular}
\end{table}

\begin{table}[ht]
  \centering
  \caption{\textbf{Diffusion Policy Hyperparameters}}
  \label{tab:diffusion_hyperparameters}
  \begin{tabular}{lr}
    \\
    \hline
    Hyperparameter & Value \\
    \hline
    \multicolumn{2}{c}{Architecture} \\
    \hline
    Horizon & 16 \\
    Observation History & 4 \\
    Embedding dimension & 512 \\
    Attention heads & 8 \\
    Transformer layers & 6 \\
    Denoising steps & 20 \\
    \hline
    \multicolumn{2}{c}{Training} \\
    \hline
    Batch size & 512 \\
    Number of epochs & 1000 \\
    Learning rate & $1 \times 10^{-4}$ \\
    Weight decay & 0.001 \\
    LR scheduler & Cosine \\
    LR warmup gradient steps & 10,000 \\
    EMA power & 0.75 \\
    EMA max value & 0.9999 \\
    \hline
  \end{tabular}
\end{table}

\begin{table}[ht]
    \centering
    \fontsize{8}{9.6}\selectfont
    \setlength{\tabcolsep}{3pt}
    \caption{\textbf{Motion Segments Tested in Sim and Real.}}
    \label{tab:skill_success}
    \begin{tabular}{lcc}
        \hline
        Name & Sim & Real [s] \\
        \hline
        \multicolumn{3}{c}{Short Sequence} \\
        \hline
        Cristiano Ronaldo~\cite{he2025asap} & Full & Full \\
        Side Kick~\cite{xie2025kungfubot} & Full & Full \\
        Single Leg Balance~\cite{zhang2025hub} & Full & Full \\
        Swallow Balance~\cite{zhang2025hub} & Full & Full \\
        Aerial Cartwheel & Full & Full \\
        Double Kicks 1 & Full & Full \\
        Double Kicks 2 & Full & Full \\

        \hline
        \multicolumn{3}{c}{LAFAN1 \cite{lafan1} (about 3 minutes each)} \\
        \hline
        walk1\_subject1 & Full & [0.0, 33.0], [81.2, 86.7] \\
        walk1\_subject2 & Full & - \\
        walk1\_subject5 & Full & [146.7, 159.0], [206.7, 263.7] \\
        walk2\_subject1 & Full & - \\
        walk2\_subject3 & Full & [42.7, 75.7], [217.6, 230.6] \\
        walk2\_subject4 & Full & [154.4, 164.4], [218.6, 238.6] \\
        dance1\_subject1 & Full & [0.0, 118.0] \\
        dance1\_subject2 & Full & Full \\
        dance1\_subject3 & Full & - \\
        dance2\_subject1 & Full & - \\
        dance2\_subject2 & Full & - \\
        dance2\_subject3 & Full & [43.1, 163.1], [164.3, 184.3] \\
        dance2\_subject4 & Full & [156.3, end] \\
        dance2\_subject4 & Full & - \\
        fallAndGetUp1\_subject4 & Full & - \\
        fallAndGetUp2\_subject2 & Full & [0.0, 21.0], [74.0, 91.2], [94.0, 109.0] \\
        fallAndGetUp2\_subject3 & Full & [26.5, 46.5] \\
        run1\_subject2 & Full & [0.0, 50.0] \\
        run1\_subject4 & Full & - \\
        run1\_subject5 & Full & - \\
        run2\_subject1 & Full & [0.0, 11.0], [167.4, 204.4] \\
        jumps1\_subject1 & Full & [24.3, 42.3], [71.6, 81.6], [205.5, 226.5] \\
        jumps1\_subject2 & Full & - \\
        jumps1\_subject5 & Full & - \\
        fightAndSports1\_subject1 & Full & [16.8, 25.4], [201.6, end] \\
        fightAndSports1\_subject4 & Full & - \\
        fight1\_subject2 & Full & - \\
        fight1\_subject3 & Full & - \\
        fight1\_subject5 & Full & - \\
        \hline
    \end{tabular}
\end{table}


\end{document}